%% file: acl_latex.tex
\pdfoutput=1

\PassOptionsToPackage{table}{xcolor}

\documentclass[11pt]{article}

\usepackage[final]{acl}

\usepackage{times}
\usepackage{latexsym}
\usepackage{multirow}
\usepackage{booktabs}
\usepackage{pifont} 
\usepackage{xspace}
\usepackage[most]{tcolorbox}
\usepackage[utf8]{inputenc}   
\usepackage{xcolor}
\usepackage{listings}

\lstdefinestyle{templcode}{
  language=Python,
  basicstyle=\ttfamily\footnotesize,
  backgroundcolor=\color{black!3},
  frame=single,
  rulecolor=\color{black!30},
  breaklines=true,
  breakatwhitespace=false,
  keepspaces=true,
  columns=fullflexible,
  showstringspaces=false,
  keywordstyle=\bfseries,
  commentstyle=\color{black!55},
  stringstyle=\color{black},
  literate=
    {×}{{$\times$}}1
    {−}{{$-$}}1
    {–}{{$-$}}1
    {÷}{{$\div$}}1
}

\newcommand{\ie}{i.e.\xspace}
\usepackage{amsmath}
\usepackage{amssymb}
\usepackage{arydshln}
\usepackage[colorinlistoftodos,textsize=scriptsize]{todonotes}
\usepackage{adjustbox} 
\usepackage[most]{tcolorbox}
\usepackage{paralist}
\usepackage[nameinlink]{cleveref}

\definecolor{aclblue}{RGB}{0,85,160}

\usepackage{array}
\newcolumntype{L}[1]{>{\raggedright\arraybackslash}p{#1}}

\newcommand{\dataset}[1]{\textsc{#1}\xspace}
\newcommand{\ourdataset}{\dataset{FinChain}}

\newcommand{\metric}[1]{\textsc{#1}\xspace}
\newcommand{\ourmetric}{\metric{ChainEval}}

\usepackage[T1]{fontenc}

\usepackage[utf8]{inputenc}
\usepackage{fontawesome5}
\usepackage{microtype}

\usepackage{inconsolata}

\usepackage{graphicx}

\title{\includegraphics[height=1.5em]{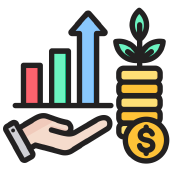}\ \ourdataset: A Symbolic Benchmark\\ for Verifiable Chain-of-Thought Financial Reasoning 
}

\author{
 \textbf{Zhuohan Xie\textsuperscript{1}} \ 
 \textbf{Daniil Orel\textsuperscript{1}}\thanks{Co-second author} \ 
 \textbf{Rushil Thareja\textsuperscript{1}}\footnotemark[1]  \
 \textbf{Dhruv Sahnan\textsuperscript{1}}\footnotemark[1] \ 
 \textbf{Hachem Madmoun\textsuperscript{1,2}} \\
 \textbf{Fan Zhang\textsuperscript{3}} \
 \textbf{Debopriyo Banerjee\textsuperscript{1}} \
 \textbf{Georgi Georgiev\textsuperscript{4}} \
 \textbf{Xueqing Peng\textsuperscript{5}}\thanks{Corresponding author} \
 \textbf{Lingfei Qian\textsuperscript{5}} \\
 \textbf{Jimin Huang\textsuperscript{5}} \
 \textbf{Jinyan Su\textsuperscript{6}} \
 \textbf{Aaryamonvikram Singh\textsuperscript{1}} \
 \textbf{Rui Xing\textsuperscript{7}} \
 \textbf{Rania Elbadry\textsuperscript{1}} \\
 \textbf{Chen Xu\textsuperscript{5}}
 \textbf{Haonan Li\textsuperscript{1}} \
 \textbf{Fajri Koto\textsuperscript{1}} \
 \textbf{Ivan Koychev\textsuperscript{4}} \
 \textbf{Tanmoy Chakraborty\textsuperscript{8}} \\
 \textbf{Yuxia Wang\textsuperscript{9}} \
 \textbf{Salem Lahlou\textsuperscript{1}} \
 \textbf{Veselin Stoyanov\textsuperscript{1}} \
 \textbf{Sophia Ananiadou\textsuperscript{10}} \ 
 \textbf{Preslav Nakov\textsuperscript{1}} \\
 \textsuperscript{1}MBZUAI \
 \textsuperscript{2}Syllogia \
 \textsuperscript{3}The University of Tokyo \
 \textsuperscript{4}Sofia University ``St. Kliment Ohridski'' \\
 \textsuperscript{5}The Fin AI \
 \textsuperscript{6}Cornell University \
 \textsuperscript{7}The University of Melbourne \
 \textsuperscript{8}IIT Delhi \\
 \textsuperscript{9}INSAIT, Sofia University ``St. Kliment Ohridski'' \
 \textsuperscript{10}The University of Manchester\\
\texttt{\{zhuohan.xie, preslav.nakov\}@mbzuai.ac.ae} \
\texttt{xueqing.peng2024@gmail.com} \\
\faGlobe\ \href{https://mbzuai-nlp.github.io/finchain/}{\textcolor{aclblue}{Project}}
\quad
\faGithub\ \href{https://github.com/mbzuai-nlp/finchain}{\textcolor{aclblue}{Code}}
\quad
\faMedal\ \href{https://mbzuai-nlp.github.io/finchain/leaderboard.html}{\textcolor{aclblue}{Leaderboard}}
}

\begin{document}
\maketitle


\input{sections/0_abstract}
\input{sections/1_intro}

\input{sections/2_relatedwork}

\input{sections/3_dataset}

\input{sections/4_metric}

\input{sections/5_experiments}


\section{Conclusion and Future Work}

We introduced \ourdataset, a symbolic benchmark for verifiable Chain-of-Thought financial reasoning, spanning 58 topics across 12 domains and three difficulty levels. To support step-level evaluation, we proposed \ourmetric, which jointly assesses intermediate reasoning consistency and final-answer correctness. Our results showed that while frontier LLMs perform best overall, even the strongest models struggle with complex symbolic reasoning, and fine-tuned open-source systems narrow down, but do not close this gap.

In future work, we plan to extend \ourdataset to multilingual and region-specific settings and to incorporate problems grounded in real-world financial documents. This direction aims to bridge symbolic reasoning and factual verification~\citep{xie-etal-2025-fire}, advancing more interpretable and reliable financial AI systems.

\section*{Limitations}

This work has several limitations, which can be addressed in future research.
First, our dataset is entirely synthetic and generated from symbolic templates. While this design enables controllable, contamination-free generation and automatic verification of both the reasoning chain and the final answer, it may lack the linguistic diversity and contextual richness of real-world financial texts. Future work could incorporate real financial documents as seed inputs for semi-structured generation while preserving symbolic grounding.



Second, the benchmark focuses on symbolic numerical reasoning and does not capture qualitative or strategic aspects of financial decision-making (e.g., risk assessment or market sentiment). Thus, \ourdataset{} does not cover the full spectrum of financial reasoning, and extending it to higher-level reasoning remains an open challenge.

Third, \ourdataset{} is limited to English and U.S.-centric financial conventions, limiting its applicability to multilingual and regional contexts. Extending it to other languages and financial systems is an important direction for future work.

Finally, our evaluation relies on parsing model-generated reasoning chains, which is sensitive to formatting variations and extraneous text. Improving step-level alignment via more structured outputs or tighter integration with symbolic execution is a promising direction for future work.


\section*{Ethical Statement and Broad Impact}


We use only synthetic data generated via templated code and language models, without any private, sensitive, or copyrighted content. The benchmark promotes transparency and reproducibility in financial AI, but caution is needed when deploying LLMs in real-world financial decision-making, especially where correctness and regulatory compliance are critical. We hope \ourdataset{} supports research toward more interpretable, verifiable, and safe reasoning systems in high-stakes domains.

\paragraph{Data License}

The \ourdataset dataset and accompanying code is released under the MIT License.

\section*{Acknowledgments}


We would like to express our sincere gratitude to our financial experts, Salim Tlemçani, Alfred Choi, Petrus Kung, Shaobo Wang, Bowen Hao, and Xunwen Zheng, for contributing their time, expertise, and careful reviews of the templates throughout the development of this benchmark. Their feedback and domain knowledge were invaluable to this project. We are also deeply grateful to our student volunteers, Muhammad Usman Safder and Ayesha Gull, for their enthusiasm and dedication in developing the live demo and website.

We further thank The Fin AI community for its research support, constructive feedback, and collaborative environment, which meaningfully contributed to this work.

Finally, we thank the anonymous reviewers for their constructive feedback.

\bibliography{custom}

\newpage

\input{sections/7_appendix}

\end{document}

%% file: sections/0_abstract.tex
\begin{abstract}

Multi-step symbolic reasoning is essential for robust financial analysis; yet, current benchmarks largely overlook this capability. Existing datasets such as FinQA and ConvFinQA emphasize final numerical answers while neglecting the intermediate reasoning steps required for transparency and verification. To address this gap, we introduce \ourdataset, the first benchmark specifically designed for verifiable Chain-of-Thought evaluation in finance. \ourdataset spans 58 topics across 12 financial domains, each represented by parameterized symbolic templates with executable Python code that enable fully machine-verifiable reasoning and scalable, contamination-free data generation.
To assess reasoning capacity, we propose \ourmetric, a dynamic alignment measure that jointly evaluates both the final-answer correctness and the step-level reasoning consistency. Our evaluation of 26 leading LLMs reveals that even frontier LLMs exhibit clear limitations in symbolic financial reasoning, while domain-adapted and math-enhanced fine-tuned models can substantially narrow this gap.
Overall, \ourdataset exposes persistent weaknesses in multi-step financial reasoning and provides a foundation for developing trustworthy, interpretable, and verifiable financial AI.
This project is available at \url{https://github.com/mbzuai-nlp/finchain.git}.

\end{abstract}

%% file: sections/1_intro.tex
\section{Introduction}
\label{sec-intro}
Large language models (LLMs) have demonstrated strong performance across a wide range of tasks~\citep{llm_survey, xie-etal-2023-next}.

These models have likewise shown promise in financial applications~\citep{chen2024survey, xie2026clef}, where effective analysis often requires synthesizing large volumes of textual information from reports, news, and social media, which reflect and influence financial phenomena such as investor sentiment, risk perceptions, and expected market trends~\citep{nie2024survey, zhang2026finreporting, zhou2026fincards}.
\begin{figure}[t]
\centering
\resizebox{\columnwidth}{!}{%
\input{figs/example}
}
\caption{\textbf{Symbolic template for generating compound interest problems in \ourdataset.}}
\label{fig:template}
\end{figure}

Most prior work in financial NLP has focused on tasks with shallow supervision, including information extraction~\citep{finer}, sentiment analysis~\citep{tweetfinsent}, and text classification~\citep{finarg}. 
These tasks typically require models to produce short outputs and do not test whether they can perform transparent, multi-step financial reasoning. 
In contrast, many financial problems require generating structured chains of reasoning that justify each intermediate step, as illustrated in~\autoref{fig:template}.
Existing financial reasoning benchmarks such as FinQA~\citep{finqa} and ConvFinQA~\citep{convfinqa} primarily frame reasoning as numerical question answering and emphasize final-answer correctness. 
While some examples include intermediate reasoning traces, these are neither systematically structured nor rigorously verified. 
As a result, such benchmarks cannot reliably diagnose where reasoning fails or distinguish genuine multi-step inference from surface-level pattern matching.

Inspired by the symbolic-template paradigm introduced in mathematical reasoning~\citep{gsmsymbolic}, we construct a financial reasoning benchmark entirely from scratch. 
As shown in~\autoref{fig:template}, each symbolic template encodes a parameterized financial problem with named variables and numeric inputs, paired with executable Python code that computes both intermediate steps and final results. 
This design supports scalable, contamination-free data generation grounded in explicit symbolic and numerical operations.
Financial reasoning spans diverse topics and requires heterogeneous expertise. 
To capture this diversity, we organize our dataset using a fine-grained taxonomy (see~\autoref{fig:taxonomy}) covering 12 domains and 58 topics. 
For each topic, we design five parameterized templates of increasing difficulty, comprising two easy, two intermediate, and one advanced template.
Each instantiated example consists of a scenario card specifying the topic, the difficulty, and some example inputs, together with an executable chain of reasoning steps grounded in domain-specific formulae. 
Because the gold reasoning traces are explicit and executable, the intermediate computations and the final results can be verified at the symbolic and the numerical levels, thus enabling automatic detection of incorrect or inconsistent reasoning steps.
To support rigorous and interpretable evaluation, we introduce \ourmetric{}, a dynamic-alignment metric that jointly evaluates final-answer correctness and intermediate step faithfulness. 

Unlike conventional text similarity measures, \ourmetric{} explicitly accounts for both semantic correspondence and numerical consistency between predicted and reference reasoning chains.
Using this benchmark and evaluation framework, we evaluate 26 proprietary and open-weight LLMs. 
We find that frontier LLMs perform best overall, yet consistently struggle with advanced multi-step symbolic financial reasoning, while fine-tuned compact models achieve only limited gains.

Our main contributions are as follows:
\begin{itemize}
\item We introduce the first from-scratch symbolic benchmark for financial reasoning, grounded in a fine-grained taxonomy spanning 12 domains and 58 topics.
\item We propose \ourmetric{}, a verifiable reasoning measure that evaluates both step-level consistency and final-answer correctness, and shows the strongest correlation with expert human judgments.
\item We benchmark 26 leading proprietary and open-weight LLMs, and find that even state-of-the-art LLMs struggle with verifiable multi-step financial reasoning, particularly on advanced symbolic templates.
\end{itemize}

%% file: figs/example.tex
\definecolor{varname}{HTML}{CDE7FF}
\definecolor{varproj}{HTML}{FFE5B4}
\definecolor{vartotal}{HTML}{C3F5D5}
\definecolor{varrate}{HTML}{FFCEDC}
\definecolor{vartime}{HTML}{FFF8B0}
\definecolor{varprincipal}{HTML}{D9CCFF}
\definecolor{varamount}{HTML}{E0F2F1}
\definecolor{varci}{HTML}{E2E2E2}
\definecolor{headergray}{HTML}{555555}
\newcommand{\dashedline}{%
  \noindent
  \begin{center}
    \begin{tikzpicture}
      \draw[dashed, gray, line width=0.5pt] (0,0) -- (6.5,0);
    \end{tikzpicture}
  \end{center}
}


\begin{tcolorbox}[title=\ourdataset (Compound Interest), colback=white, colframe=headergray, boxrule=0.5pt, arc=1mm, fonttitle=\bfseries]

\noindent \textbf{\#Question:} \\
\colorbox{varname}{\texttt{investor\_name}} invested \colorbox{varprincipal}{\texttt{principal}} in \colorbox{varproj}{\texttt{project\_name}}.  
The investment grows at an annual interest rate of \colorbox{varrate}{\texttt{rate}}\% compounded annually over \colorbox{vartime}{\texttt{time}} years.  
Calculate the \colorbox{varci}{\texttt{compound interest (CI)}}.

\vspace{0.5em}

\noindent \textbf{\#Variables:} \\
- \colorbox{varname}{\texttt{investor\_name}} = sample(investors) \\
- \colorbox{varproj}{\texttt{project\_name}} = sample(projects) \\
- \colorbox{varprincipal}{\texttt{principal}} = range(1000, 5000) \\
- \colorbox{varrate}{\texttt{rate}} = uniform(2, 10) \\
- \colorbox{vartime}{\texttt{time}} = range(1, 5)

\dashedline

\noindent \textbf{\#Chain-of-Thought Solution:} \\
Step 1: Compute the compound amount:
\colorbox{varamount}{\texttt{amount}} = \colorbox{varprincipal}{\texttt{principal}} $\times$ $\left(1 + \frac{\colorbox{varrate}{\texttt{rate}}}{100}\right)^{\colorbox{vartime}{\texttt{time}}}$ \\
Step 2: Compute the compound interest:
  \quad \colorbox{varci}{\texttt{CI}} = \colorbox{varamount}{\texttt{amount}} $-$ \colorbox{varprincipal}{\texttt{P}}

\end{tcolorbox}

%% file: sections/2_relatedwork.tex
\section{Related Work}

\begin{figure*}[t]
    \centering
    \includegraphics[width=0.98\linewidth]{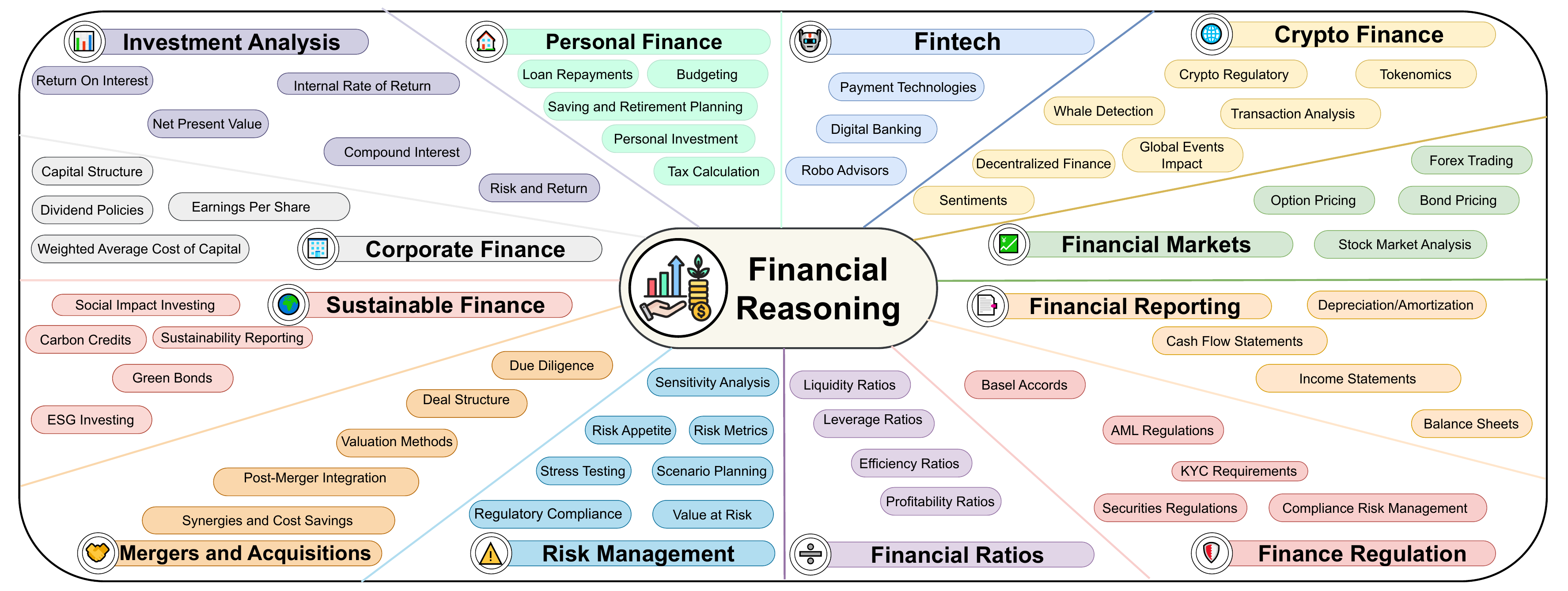}
    \caption{
\textbf{\ourdataset taxonomy of financial reasoning topics.}
Our benchmark spans 58 topics organized into 12 major domains, ranging from traditional areas like \textit{Corporate Finance} and \textit{Financial Reporting} to emerging fields such as \textit{Crypto Finance} and \textit{Sustainable Finance}. This hierarchical structure enables fine-grained evaluation of symbolic reasoning across diverse financial domains.
}
    \label{fig:taxonomy}
\end{figure*}

\subsection{Financial NLP}

Progress in financial NLP has been driven by both modeling and benchmarking. Early work focused on extraction and classification with models such as FinBERT~\citep{araci2019finbert}, while later efforts expanded to personal finance~\citep{aipersonalfinance}, credit scoring~\citep{generalistcreditscoring}, and risk-awareness benchmarking~\citep{yuan-etal-2024-r}. Datasets like FiNER-ORD, REFinD, FinARG, and ECTSum support tasks in NER, relation extraction, argument mining, and summarization~\citep{finer, kaur2023REFinD, ectsum, finben}.
Large financial language models have further advanced the field. BloombergGPT~\citep{wu2023bloomberggpt} achieved broad in-domain performance, FinGPT~\citep{liu2023fingpt} emphasized open-source adaptability, and FinMA~\citep{pixiu} delivered competitive results with a compact architecture. Corresponding benchmarks such as FLANG~\citep{shah-etal-2022-flue}, FinBen~\citep{finben}, and FinMTEB~\citep{finmteb} broadened evaluation coverage across diverse tasks.

Furthermore, BizBench~\citep{bizbench} and PIXIU~\citep{pixiu} evaluated LLMs on quantitative and multimodal reasoning.
Despite this progress, limitations remain in multi-step reasoning, long-context understanding, and cross-market generalization~\citep{chen2024survey}.
These gaps motivate the need for benchmarks that assess the capability of LLMs to perform faithful, audit-able reasoning grounded in financial knowledge.

\subsection{Financial Reasoning}

Real-world financial problems require precise numerical reasoning.
FinQA~\citep{finqa} and ConvFinQA~\citep{convfinqa} were developed before Chain-of-Thought reasoning became a standard evaluation target. As a result, they supervise arithmetic program generation but offer only weak step-level signals, yielding traces that are neither explicit nor verifiable.
FinTextQA~\citep{fintextqa} introduces long-form financial questions from textbooks and regulatory sources and focuses on explanatory retrieval rather than traceable computation.
Bridging text and numerical reasoning, TAT-QA~\citep{tatqa} and MultiHiertt~\citep{multihiertt} combine textual and tabular evidence, while DocMath-Eval~\citep{docmath} and FinanceMath~\citep{financemath} move toward interpretable symbolic evaluation. However, these datasets remain largely domain-agnostic and lack explicit, step-level supervision grounded in financial formulae.
More recently, FinanceReasoning~\citep{financereasoning} improves answer-level numerical reliability by introducing executable Python solutions.

However, FinanceReasonin does not provide systematic verification of step-level reasoning alignment.
\ourdataset addresses this gap by introducing a symbolic, executable benchmark with explicit intermediate supervision and automatic alignment-based evaluation, spanning 58 topics across 12 financial domains.

%% file: sections/3_dataset.tex
\section{\ourdataset}
\label{finchain}


\subsection{Data Creation Process}

We begin by identifying and defining financial domains based on established literature~\citep{finance_textbook25} and expert input within the team, resulting in 12 distinct domains.
Within each domain, we propose candidate financial topics with LLM assistance and curate them with financial experts, yielding a total of 58 topics (mean 4.8 per domain). The resulting taxonomy is illustrated in \autoref{fig:taxonomy}.
Following \citet{gsmsymbolic}, we instantiate each topic through parameterized symbolic templates that define both the question structure and an executable Chain-of-Thought solution grounded in domain-specific formulas. We implement these templates as executable Python programs that generate both intermediate reasoning steps and final answers, thereby enabling fully machine-verifiable evaluation.
For each topic, we design five templates spanning three difficulty levels (two basic, two intermediate, and one advanced), where we control difficulty by the number and the complexity of required reasoning steps.
\autoref{tab:dataset-stats} summarizes the dataset statistics and shows increasing reasoning depth across difficulty levels.

An example symbolic template is shown in \autoref{fig:template}.
We generate the templates with LLM assistance and subsequently curate them with domain experts to ensure correctness, consistency, and balanced difficulty. We provide detailed prompt designs and generation procedures in Appendix~\ref{templatecreation}.
This design isolates financial reasoning ability from document parsing challenges. We therefore position \ourdataset\ as a controlled testbed for verifiable financial reasoning, complementary to benchmarks centered on real-world document understanding.

\begin{table}[t]
\centering
\small
\begin{tabular}{lccc}
\toprule
\textbf{Statistic} & \textbf{Basic} & \textbf{Intermediate} & \textbf{Advanced} \\
\midrule
\#Templates & 116 & 116 & 58 \\
Avg. steps & 2.01 & 2.97 & 3.90 \\
\bottomrule
\end{tabular}
\caption{Dataset statistics of \ourdataset.}
\label{tab:dataset-stats}
\end{table}

\subsection{Data Validation and Expert Review}

To ensure data quality and consistency, we apply a set of validation constraints covering numerical precision, unit consistency, input completeness, and reasoning step informativeness. Templates that fail validation are revised prior to expert review. A detailed description of the validation criteria is provided in Appendix~\ref{validation}.
We ask financial experts to review all validated templates following a calibrated annotation protocol. Specifically, all reviewers first participated in a pilot calibration phase, during which they jointly reviewed a shared subset of templates and discussed discrepancies to align on annotation standards. After this calibration phase, reviewers independently assessed the remaining templates, evaluating both the correctness of reasoning steps and the final numerical results under the agreed-upon criteria. Further details on annotator backgrounds, annotation procedures, and quality control are provided in Appendix~\ref{expertreview}.

%% file: sections/4_metric.tex
\section{\ourmetric}


We propose \ourmetric, an evaluation framework that jointly assesses reasoning-step alignment and final-answer correctness. Building on prior work on reasoning consistency~\citep{faithfulcot, roscoe}, our approach explicitly verifies intermediate results via step–answer matching while also checking the final numerical outcome.

\subsection{Preliminaries}
We define the gold solution $S^{*}$ and the predicted solution $\hat{S}$ as sequences of $m$ and $n$ reasoning steps, respectively:
\begin{align}
    S^* = (s^*_1, \ldots, s^*_m), 
    \quad
    \hat{S} = (\hat{s}_1, \ldots, \hat{s}_n),
\end{align}
where $s^*_i$ and $\hat{s}_j$ denote individual reasoning steps in the gold and in the predicted solutions, respectively.
Each step $s_i$ produces an intermediate result,
\begin{equation}
    \mathrm{StepRes}(s_i) = a_i,
\end{equation}
representing the numerical or the symbolic value computed at that step.

To evaluate the reasoning faithfulness, we compare these sequences both semantically and numerically.  
In addition, we apply \emph{Dynamic Time Warping} (DTW) to capture the global structural alignment between step sequences.  
DTW provides an order-preserving but flexible alignment that accommodates insertions, deletions, or small reordering of steps while maintaining the overall sequence coherence.

\subsection{Reasoning Step Alignment}
We assess the consistency between gold and predicted reasoning traces through two complementary components: semantic similarity and answer-level agreement, combined within a DTW-based alignment framework.

\paragraph{Semantic Similarity.}
Each step is encoded using a sentence encoder $\mathrm{Enc}(\cdot)$, and the pairwise semantic similarity between the gold and the predicted steps is computed as
\begin{equation}
    \label{eqn:sem_sim}
    \mathrm{SS}(s^*_i, \hat{s}_j) = 
    \mathrm{cos}\!\left(\mathrm{Enc}(s^*_i), \mathrm{Enc}(\hat{s}_j)\right),
\end{equation}
where $\mathrm{cos}(\cdot,\cdot)$ denotes the cosine similarity and $\mathrm{SS} \in [0,1]$.

\paragraph{Answer Match.}
For the intermediate results produced by each step, we evaluate numeric or symbolic consistency:
\begin{align*}
    \mathrm{StepRes}(s^*_i) = a^*_i, \qquad 
    \mathrm{StepRes}(\hat{s}_j) = \hat{a}_j.
\end{align*}
We then define the answer-matching function:
\begin{equation}
\label{eqn:ans_match}
\mathrm{AM}(s^*_i, \hat{s}_j) =
\begin{cases}
  \mathbb{I}\!\left(
  \frac{|\hat{a}_j - a^*_i|}{|a^*_i|}
  \le \epsilon
  \right), \\[-2pt]
  \hspace{0.5em} \text{if both are numeric},\\[6pt]
  \mathbb{I}(\hat{a}_j = a^*_i),\ \text{otherwise.}
\end{cases}
\end{equation}

Here, $\mathbb{I}(\cdot)$ denotes the indicator function, and $\epsilon = 0.05$ permits up to a 5\% relative numerical deviation to account for rounding or error propagation.
This design choice is motivated by financial auditing standards\footnote{\url{https://www.materialitytracker.net/standards/financial-thresholds/}}, in which materiality thresholds are commonly defined as between 5\% and 10\% of a base metric such as earnings, and deviations below 5\% are generally considered immaterial.

\paragraph{Gated Step-Level Similarity.}
To ensure that a pair of steps is considered consistent only when both their semantics and results agree, we define a gated score:
\begin{equation}
    \label{eqn:gate_score}
    \mathrm{Score}_{\text{gate}}(i,j) =
    \mathrm{SS}(s^*_i, \hat{s}_j)\;
    \times\;
    \mathrm{AM}(s^*_i, \hat{s}_j).
\end{equation}
This score forms the basis of the DTW alignment matrix.

\paragraph{Dynamic Sequence Alignment.}
To capture global reasoning consistency, we align the two step sequences using 
\emph{Dynamic Time Warping} (DTW).  
DTW searches for an optimal monotonic path between $(S^*, \hat{S})$ that minimizes cumulative cost while preserving step order. This formulation allows local insertions, deletions, and compressions, and does not require strict template matching as long as the predicted reasoning remains semantically and numerically aligned with the reference.

\paragraph{DTWGate Alignment Score.}
We transform the minimal DTW cost into a normalized similarity measure as follows:
\begin{equation}
    \label{eqn:dtw_norm_gate}
    \mathrm{DTWGate}(S^*, \hat{S}) =
    1 - \frac{\mathrm{Cost}_{\mathrm{DTW}}}{L_{\mathrm{path}}},
\end{equation}
where $\mathrm{Cost}_{\mathrm{DTW}}$ denotes the total alignment cost and $L_{\mathrm{path}}$ represents the length of the optimal alignment path.  
The resulting score lies in the range $[0,1]$, with higher values indicating stronger reasoning alignment between the gold and the predicted solutions.

\subsection{Final Answer Correctness}
Beyond step-level reasoning alignment, we also assess the correctness of the final predicted outcome.
Let $s^*_m$ and $\hat{s}_n$ denote the last steps of the gold and the predicted solutions, respectively.  
We define the \textbf{Final Answer Correctness (FAC)} metric as
\begin{equation}
    \label{eqn:fac}
    \mathrm{FAC}(S^*, \hat{S}) =
    \begin{cases}
        \mathbb{I}\!\left(\frac{|\hat{a}_n - a^*_m|}{|a^*_m|} \leq \epsilon\right), \\
        \hspace{0.5em} \text{if both are numeric},\\[4pt]
        \mathbb{I}(\hat{a}_n = a^*_m), \text{otherwise},
    \end{cases}
\end{equation}

Here, we use the same tolerance $\epsilon = 0.05$ as before.
FAC measures whether the model’s final computation aligns with the correct end result, complementing the DTW-based metric that evaluates reasoning faithfulness throughout the entire solution sequence. Therefore, we have
\begin{equation}
\ourmetric
= (1 - \alpha)\cdot \text{DTWGate}
+ \alpha \cdot \text{FAC}, 
\end{equation}
which accounts for both reasoning correctness and final answer correctness. 
We set $\alpha = 0.1$, selected via grid search on a subset to maximize the correlation with human evaluations, as explained in \autoref{appx:metric_ablation}. We further verify that the final measure best reflects true reasoning quality by comparing it with human evaluations, as described in \autoref{sec:metric_validation}.

%% file: sections/5_experiments.tex
\section{Experiments and Results}

\subsection{Evaluated Models}
We evaluate a total of 26 LLMs, grouped into four categories according to their capability and relevance to financial reasoning.
(1) \textbf{Frontier proprietary models}, including \texttt{GPT-\{5, 4.1, 5-mini, 4.1-mini\}}~\citep{gpt5,gpt41}, \texttt{Claude Sonnet \{4.5, 4, 3.7\}}~\citep{claude45,claude4,claude37}, \texttt{Gemini-2.5 \{Pro, Flash\}}~\citep{gemini25}, \texttt{DeepSeek-\{V3.2, V3.1, R1\}}~\citep{liu2024deepseek,guo2025deepseek}, and \texttt{Grok-4 \{Heavy, Fast\}}~\citep{grok4}.
(2) \textbf{Finance-specific models}, including \texttt{Fin-o1}~\citep{fino1}, \texttt{Fin-R1}~\citep{finr1}, \texttt{DianJin-R1}~\citep{dianjin-r1}, and \texttt{WiroAI Finance-\{LLaMA, Qwen\}}~\citep{WiroAI-finance-qwen}.
(3) \textbf{Math-enhanced models}, including \texttt{WizardMath}~\citep{wizardmath}, \texttt{MetaMath}~\citep{metamath}, \texttt{Mathstral}~\citep{mathstral}, and \texttt{Qwen-2.5-Math}~\citep{qwen25math}.
(4)~\textbf{General-purpose open-weight models}, including \texttt{LLaMA-3.1}~\citep{llama3} and \texttt{Qwen-\{2.5, 3\}}~\citep{qwen25,qwen3}.
Detailed configurations and model sources are described in \autoref{modelcards}.

\begin{table*}[t]
\small
\centering
\renewcommand{\arraystretch}{1.09}
\setlength{\tabcolsep}{6pt}

\begin{tabular}{l c c c c c c}
\toprule
\textbf{Model} & \textbf{Size} & \textbf{\ourmetric $\uparrow$} & \textbf{FAC $\uparrow$} & \textbf{ROUGE R$_2$ $\uparrow$} & \textbf{ROUGE R$_L$ $\uparrow$} & \textbf{BERTScore $\uparrow$} \\
\midrule
\rowcolor{gray!10}
\multicolumn{7}{l}{\textbf{Frontier Proprietary LLMs}} \\
GPT-5 & N/A & 66.57$^{10.64}$ & 82.03$^{32.40}$ & \textbf{28.84}$^{12.30}$ & \textbf{42.77}$^{12.91}$ & \textbf{88.77}$^{2.39}$ \\
GPT-4.1 & N/A & 65.34$^{9.36}$ & \textbf{84.66}$^{29.26}$ & 19.38$^{8.91}$ & 30.12$^{10.57}$ & 86.04$^{2.09}$ \\
GPT-5-mini & N/A & \textbf{67.17}$^{11.06}$ & 80.28$^{32.34}$ & 26.48$^{11.99}$ & 39.74$^{12.83}$ & 88.18$^{2.35}$ \\
GPT-4.1-mini & N/A & 65.06$^{8.88}$ & 84.59$^{30.23}$ & 18.67$^{8.86}$ & 29.05$^{10.51}$ & 86.05$^{2.09}$ \\
Claude Sonnet 4.5 & N/A & 66.33$^{9.44}$ & 83.34$^{31.79}$ & 19.69$^{7.77}$ & 29.37$^{8.81}$ & 86.07$^{2.00}$ \\
Claude Sonnet 4 & N/A & 66.20$^{9.66}$ & 82.62$^{31.96}$ & 19.87$^{7.82}$ & 29.50$^{8.59}$ & 86.38$^{1.83}$ \\
Claude Sonnet 3.7 & N/A & 65.51$^{9.30}$ & 83.14$^{31.00}$ & 19.49$^{7.77}$ & 29.36$^{8.56}$ & 86.38$^{1.82}$ \\
Gemini-2.5 Pro & N/A & 66.04$^{9.71}$ & 84.34$^{30.99}$ & 17.61$^{7.07}$ & 27.36$^{8.28}$ & 85.94$^{1.88}$ \\
Gemini-2.5 Flash & N/A & 65.96$^{10.10}$ & 83.90$^{30.61}$ & 18.98$^{8.05}$ & 29.22$^{9.32}$ & 86.34$^{2.02}$ \\
DeepSeek-v3.2 & N/A & 65.23$^{9.63}$ & 84.17$^{31.23}$ & 21.73$^{10.32}$ & 32.85$^{11.31}$ & 86.66$^{2.15}$ \\
DeepSeek-v3.1 & N/A & 65.29$^{9.62}$ & 84.34$^{31.02}$ & 21.72$^{10.29}$ & 32.87$^{11.24}$ & 86.68$^{2.14}$ \\
DeepSeek-R1 & N/A & 51.22$^{11.04}$ & 28.97$^{30.29}$ & 8.67$^{7.21}$ & 12.93$^{9.72}$ & 84.39$^{1.79}$ \\
Grok-4 Fast & N/A & 60.69$^{16.26}$ & 66.54$^{42.54}$ & 21.33$^{11.09}$ & 32.25$^{13.30}$ & 86.83$^{3.06}$ \\
\rowcolor{gray!10}
\multicolumn{7}{l}{\textbf{Finance Specific LLMs}} \\
Fin-o1 & 8B & 41.50$^{12.49}$ & \textbf{52.79}$^{27.89}$ & 3.47$^{1.55}$ & 6.35$^{2.32}$ & 83.55$^{1.50}$ \\
Fin-R1 & 7B & \textbf{58.14}$^{7.32}$ & 52.76$^{28.04}$ & 5.70$^{2.44}$ & 9.22$^{3.33}$ & \textbf{84.30}$^{1.34}$ \\
DianJin-R1 & 7B & 51.95$^{8.98}$ & 37.69$^{23.73}$ & 6.28$^{2.95}$ & 10.79$^{4.19}$ & 83.12$^{1.32}$ \\
Finance-LLaMA & 8B & 41.35$^{10.49}$ & 25.21$^{25.64}$ & 9.39$^{4.69}$ & 16.19$^{5.84}$ & 83.48$^{2.09}$ \\
Finance-Qwen & 7B & 34.57$^{11.01}$ & 31.62$^{25.62}$ & \textbf{9.50}$^{4.26}$ & \textbf{16.46}$^{5.44}$ & 83.35$^{1.70}$ \\
\rowcolor{gray!10}
\multicolumn{7}{l}{\textbf{Math Enhanced LLMs}} \\
WizardMath & 7B & 24.33$^{15.00}$ & 41.28$^{35.69}$ & 11.66$^{6.57}$ & 20.72$^{7.83}$ & 84.78$^{2.36}$ \\
MetaMath & 7B & 7.93$^{9.43}$ & 23.97$^{28.76}$ & 11.45$^{7.36}$ & 21.08$^{9.24}$ & 84.86$^{2.99}$ \\
Mathstral & 7B & \textbf{59.87}$^{10.02}$ & 54.03$^{36.61}$ & \textbf{16.79}$^{7.82}$ & \textbf{26.97}$^{9.34}$ & \textbf{86.13}$^{2.18}$ \\
Qwen-2.5-Math & 7B & 55.35$^{14.98}$ & \textbf{62.62}$^{34.56}$ & 11.74$^{5.87}$ & 20.56$^{7.61}$ & 83.45$^{1.85}$ \\
\rowcolor{gray!10}
\multicolumn{7}{l}{\textbf{General Purpose Open LLMs}} \\
LLaMA-3.1 Instruct & 8B & 53.99$^{6.02}$ & 32.72$^{27.46}$ & 4.61$^{2.28}$ & 8.09$^{3.02}$ & 83.35$^{1.36}$ \\
Qwen-2.5 Instruct & 7B & \textbf{60.35}$^{7.47}$ & \textbf{65.41}$^{32.53}$ & \textbf{9.20}$^{4.51}$ & \textbf{15.26}$^{5.85}$ & \textbf{84.22}$^{1.78}$ \\
Qwen-3 & 8B & 43.32$^{11.81}$ & 32.28$^{28.58}$ & 4.05$^{1.69}$ & 6.61$^{2.14}$ & 83.58$^{1.24}$ \\
\bottomrule
\end{tabular}
\caption{
\textbf{Zero-shot performance across financial, mathematical, and general reasoning benchmarks.}
Scores are reported as percentages, with standard deviation in superscript. Model size (\textit{N/A}) denotes proprietary or undisclosed configurations. Within each model group, the best-performing system for each metric is highlighted in bold.
}
\label{tab:overall_model_performance}
\end{table*}

\subsection{\ourmetric Validation}
\label{sec:metric_validation}

Before conducting large-scale experiments, we validated the proposed \ourmetric metric through a controlled expert evaluation. We randomly sampled 20 instantiated questions from the \ourdataset dataset and generated answers using five models of different capacities and training paradigms, namely \texttt{GPT-5}, \texttt{GPT-4.1 mini}, \texttt{MetaMath}, \texttt{Fin-o1}, and \texttt{LLaMA-3.1}, to ensure a clear range of reasoning quality, producing 100 model-generated responses.

These responses were then randomly shuffled and anonymized for human assessment.
Financial experts independently evaluated each response with respect to the corresponding question and gold-standard reasoning trace.
They rated each output along two dimensions, \textit{Reasoning Process Quality} and \textit{Final Answer Accuracy}, on a five-point scale as described in Appendix~\ref{rubrics}.
We observed a strong association between these dimensions, with Spearman’s $\rho$ exceeding 0.94, showing that coherent reasoning often leads to accurate final answers. Consequently, subsequent analysis below will focus on \textit{Reasoning Process Quality} as the primary evaluation dimension. We further computed several variants of \ourmetric\ and compared their correlations with expert process scores against reference-based evaluation measures such as ROUGE-2 and ROUGE-L~\citep{rouge}. Among all variants, the DTWNormGate formulation with inclusion of final answer's correctness showed the highest correlation with human judgments, capturing both semantic and numeric consistency across reasoning steps. 

Therefore, we adopted it as the primary evaluation measure in this study, as detailed ablation and sensitivity analyses in \autoref{appx:metric_ablation} show that it achieves the strongest correlation with human judgments and remains stable across a broad range of $\epsilon$ values, with the best result at $\epsilon = 0.05$.

\subsection{Experimental Setup}



We instantiate the \ourdataset benchmark by sampling 10 instances per symbolic template with distinct random seeds, yielding 
58 topics $\times$ 5 templates $\times$ 10 instances = 2,900 test cases.
We evaluate all models under a unified decoding configuration:
temperature = 0.7, top-\textit{p} = 0.95, and a maximum token limit of 4,096 unless these parameters are unavailable for a given model.
We use a zero-shot setup with a standardized reasoning prompt:
\begin{quote}
\small
\begin{verbatim}
Please answer the given question and provide a
step-by-step solution.
Use the format: Step 1: ..., Step 2: ..., ...
The question is: {q}
\end{verbatim}
\end{quote}

We use \ourmetric\ as the primary evaluation measure, as it jointly measures final-answer correctness and alignment of the intermediate reasoning steps.
We post-processed model outputs with regular expressions to extract the ordered list of reasoning steps, accommodating common variations such as ``Step x:'', ``Step x'', or ``stepx''. For comparison, we also report results using frequently used reference-based measures~\citep{deltascore} to assess surface-level quality, including ROUGE-2 and ROUGE-L and BERTScore~\citep{bertscore}.

\subsection{Results}


\subsubsection{Overall Model Performance}
\label{mainresults}

As shown in \autoref{tab:overall_model_performance}, frontier proprietary models achieve the strongest overall performance under \ourmetric{}, indicating higher alignment quality in step-level symbolic reasoning. The performance differences between frontier models are not strictly monotonic. For example, GPT-5-mini slightly outperforms GPT-5, despite comparable or lower scores on surface-level metrics.
Among open-source systems, Fin-R1 and Mathstral achieve \ourmetric{} scores of 58.14 and 59.87 respectively, approaching frontier-level alignment despite their substantially smaller scale. These results suggest that targeted fine-tuning and symbolic supervision can substantially improve multi-step reasoning fidelity beyond what model size alone provides. However, the effectiveness of fine-tuning varies with adaptation scope: models trained on broad mathematical corpora (e.g., Mathstral) exhibit stronger generalization across reasoning styles, whereas models tuned narrowly for finance (e.g., Finance-LLaMA and Finance-Qwen) display more variable alignment quality. Notably, the gap between FAC and \ourmetric{} for some reasoning-oriented models suggests that correct final answers do not necessarily imply faithful intermediate reasoning.
We also evaluate Grok-4~Heavy, which was among the strongest available systems at the time of evaluation. Due to its high inference cost (approximately \$0.5 per sample),
we only assessed this system on a randomly sampled subset rather than the full benchmark, with results reported in Appendix~\ref{appendix_overall}.
Overall, these results show that while frontier models lead in verifiable symbolic reasoning, model scale alone is insufficient, and structured supervision plays a critical role in achieving high step-level alignment.

\subsubsection{Performance Across Domains}

\begin{figure}[t]
    \centering
    \includegraphics[width=1\columnwidth]{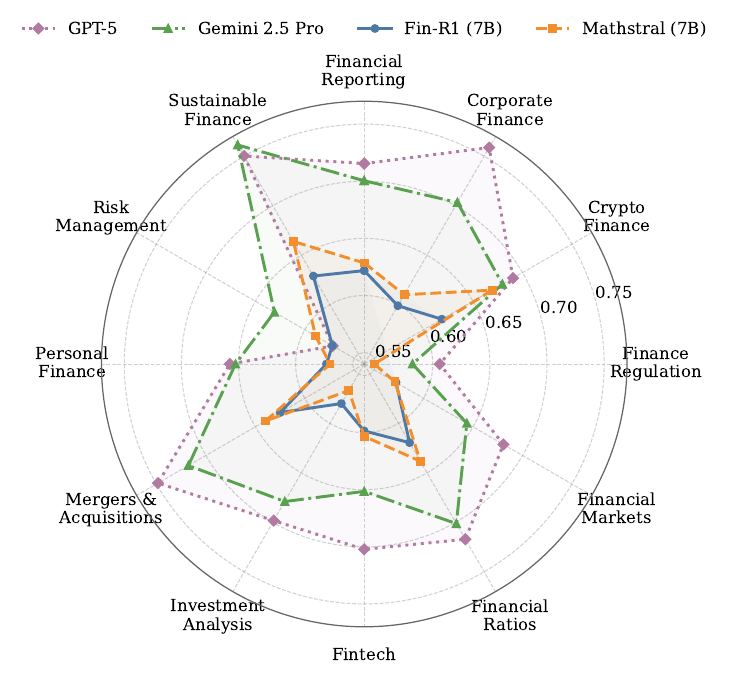}  
    \caption{
    \textbf{Domain-level performance across financial domains.}
Radar plot showing \ourmetric\ scores across twelve financial domains for four representative models: GPT-5, Gemini-2.5~Pro, Fin-R1, and Mathstral.
}
    \label{fig:topic_performance}
\end{figure}

\autoref{fig:topic_performance} shows the domain-level performance 
for four representative models: GPT-5, Gemini-2.5 Pro, Fin-R1, and Mathstral. Among them, GPT-5 demonstrates consistently strong performance across most domains, forming a relatively stable upper envelope with limited variation. Gemini-2.5 Pro follows a similar trend, achieving competitive scores across domains while exhibiting moderate domain-level differences.
The two open-weight models show more heterogeneous performance patterns. Fin-R1 achieves relatively higher scores in domains such as Financial Reporting, Sustainable Finance, and Risk Management, while exhibiting lower performance in several quantitatively intensive or structurally complex domains. In contrast, Mathstral performs competitively in quantitatively orientated domains, such as Financial Ratios and Investment Analysis, but shows weaker performance in domains with heavier textual or regulatory components.
In general, domain-level performance is not uniform across models and relative rankings vary by financial category. These observations highlight systematic variation across domains and motivate further diagnostic analysis in subsequent sections.
Domain-level results for the remaining models are provided in Appendix~\ref{appendix_domains}.
Overall, these results indicate that symbolic financial reasoning ability is highly domain-dependent, and that strong aggregate performance does not guarantee robust reasoning across diverse financial categories.

\subsubsection{Performance Across Difficulty Levels}

\begin{figure}[t]
    \centering
    \includegraphics[width=1.0\linewidth]{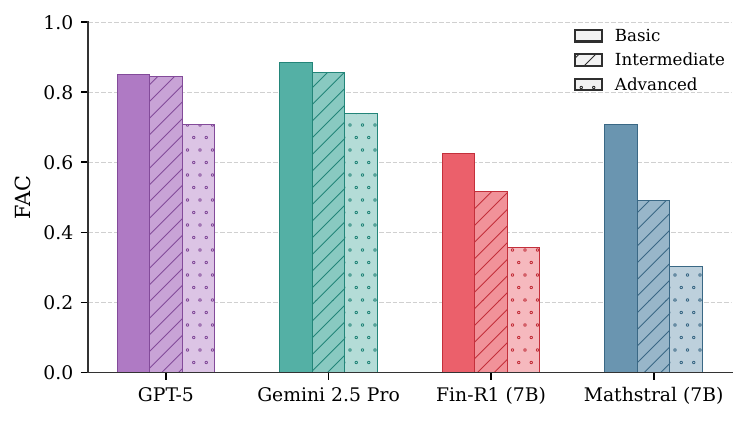}
\caption{
\textbf{Final answer correctness (FAC) across difficulty levels.}
FAC for representative models on \textit{Basic}, \textit{Intermediate}, and \textit{Advanced} \ourdataset instances.
}
    \label{fig:difficulty_performance}
\end{figure}

To assess the model robustness under increasing reasoning complexity, we group \ourdataset instances into three predefined difficulty tiers: \textit{Basic}, \textit{Intermediate}, and \textit{Advanced} (\autoref{finchain}). Each tier corresponds to an increase in the number of required reasoning steps and the depth of symbolic and numerical operations.
\autoref{fig:difficulty_performance} reports model performance across difficulty levels using final answer correctness (FAC), isolating end-task success from partial step-level alignment captured by \ourmetric{}. Across all tiers, frontier proprietary models achieve the highest correctness, with GPT-5 and Gemini-2.5~Pro maintaining relatively strong performance as task difficulty increases. Nevertheless, even these models exhibit a clear degradation on advanced instances, highlighting the challenge of solving complex, multi-step financial reasoning problems to completion.
In contrast, open-weight and fine-tuned models show substantially steeper declines as difficulty increases. Mathstral performs competitively on \textit{Basic} and \textit{Intermediate} tasks, suggesting that mathematical fine-tuning improves structured numerical reasoning, but its performance drops markedly on \textit{Advanced} instances. Fin-R1 displays similar behavior, achieving reasonable accuracy on simpler finance-oriented queries while degrading more sharply on tasks requiring longer reasoning chains.
Overall, FAC reveals a pronounced gap between frontier and non-frontier models on advanced symbolic reasoning tasks. Taken together with the more gradual degradation observed under \ourmetric{}, these results suggest that completing long, multi-step reasoning chains remains a key bottleneck.
We report the corresponding difficulty breakdown under \ourmetric{} in Appendix~\ref{app_diff_chaineval}.

\subsection{Error Analysis}
\label{sec:error_analysis}

\begin{figure}[t]
    \centering
    \includegraphics[width=1.0\columnwidth]{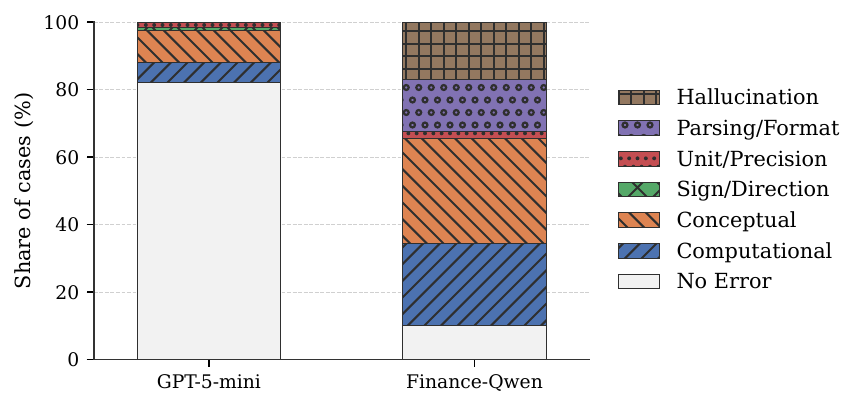}
    \caption{
\textbf{Error type distribution on a randomly sampled subset of 200 questions.}
The figure shows the distributions for GPT-5-mini and Finance-Qwen; the expanded comparison is provided in Appendix~\ref{appx:error_analysis_extended}.
}
    \label{fig:error_distribution}
\end{figure}

We conduct a targeted error analysis on a randomly sampled subset of 200 questions from \ourdataset. In \autoref{fig:error_distribution}, we analyze the same question set for GPT-5-mini and Finance-Qwen, which serve as contrasting points on the performance spectrum under an identical evaluation setup, with GPT-5-mini as a strong frontier-aligned reference and Finance-Qwen as a finance-tuned model with lower overall accuracy. This shared subset enables a direct comparison of error profiles without confounding effects from domain coverage or evaluation scale. We further extend the same analysis to six models in Appendix~\ref{appx:error_analysis_extended}, where we observe the same qualitative patterns across the expanded comparison.
We manually inspect model outputs and assign them to a small set of coarse-grained error categories capturing common reasoning failures in financial problem solving. Our taxonomy includes \textit{Computational} errors (incorrect numerical execution), \textit{Conceptual} errors (incorrect application of financial concepts), \textit{Sign/Direction} errors, \textit{Unit/Precision} mismatches, \textit{Parsing/Format} issues that prevent reliable step alignment, and \textit{Hallucination}, which includes content unsupported by the input. Outputs that are fully correct within tolerance are labeled as \textit{No Error}. A detailed description of each category is provided in Appendix~\ref{appendix_error_taxonomy}. During expert review, a small number of borderline cases initially categorized as computational errors were revised to \textit{No Error} after manual inspection, primarily when numerical deviations were deemed immaterial. All reported statistics reflect these expert-validated labels.
\autoref{fig:error_distribution} summarizes the resulting error distributions for GPT-5-mini and Finance-Qwen. GPT-5-mini produces correct solutions for most cases, with remaining failures concentrated in numerical computation and conceptual reasoning.

Other error types are rare, and no hallucinated or unparsable outputs are observed on this subset. In contrast, Finance-Qwen exhibits errors on most evaluated cases, with failures spread across multiple categories. Conceptual and computational errors dominate, while a substantial fraction involves formatting issues or hallucinated information. Compared to GPT-5-mini, Finance-Qwen errors are less concentrated in a single mode and reflect more diverse failure behaviors. 
Examples of representative errors are provided in Appendix~\ref{appendix_error_examples}, and the expanded six-model comparison is reported in Appendix~\ref{appx:error_analysis_extended}. Overall, this analysis shows that symbolic financial reasoning errors are heterogeneous and model-dependent, with weaker models showing more diverse and compounding failure modes, which explains the aggregate performance gaps.

%% file: sections/7_appendix.tex
\appendix

\section{\ourdataset Construction Details}

\subsection{Template Creation Prompt}
\label{templatecreation}

To construct symbolic financial reasoning benchmarks analogous to GSM-Symbolic, we design a structured prompt that guides the generation of executable financial templates. These templates support variable-based instantiation, symbolic step-wise supervision, and controlled perturbations for robustness evaluation. Below, we present the prompt used for template construction.
\\
\textbf{System Instruction:}
You are a financial NLP expert developing symbolic reasoning datasets. Your task is to construct symbolic templates for financial reasoning problems. Each template should support (i) controlled generation of diverse question instances, (ii) executable reasoning traces for automatic verification, and (iii) systematic variation in surface and logical complexity.

Please follow the steps below:

\begin{enumerate}
    \item \textbf{Identify a financial reasoning task:} For example, compound interest, IRR, ROI, NPV, breakeven analysis, loan amortization, etc.
    
    \item \textbf{Write a natural language question template:} Formulate the question using variable placeholders instead of fixed values. For instance, use \texttt{\{principal\}}, \texttt{\{rate\}}, \texttt{\{years\}}, etc.
    
    \item \textbf{Define variables and constraints:} Specify the domain (e.g., numerical range or categorical values) for each variable. Add algebraic constraints to ensure the question is solvable and the answer valid.
    
    \item \textbf{Write a symbolic solution trace:} Provide a step-by-step solution using the variables. Ensure the reasoning chain is executable in Python for automatic evaluation.
    
    \item \textbf{Vary difficulty levels:} For each task, generate 10 templates with increasing complexity. Longer and more compositional reasoning chains should correspond to harder levels.
\end{enumerate}

\subsection{Template Examples}
\label{templateexamples}

Here, we present example templates of three compound interest (CI) financial questions, grouped by difficulty level, including basic, intermediate, and advanced.

\paragraph{Basic Level}\mbox{}\\[-0.5em]
\input{tables/example1}

\newpage
\paragraph{Intermediate Level}\mbox{}\\[-0.5em]
\input{tables/example2}

\paragraph{Advanced Level}\mbox{}\\[-0.5em]
\input{tables/example3}

\subsection{Data Validation Criteria}
\label{validation}

Directly prompting large language models to generate symbolic financial reasoning templates can lead to inconsistencies or incomplete specifications. To address these issues, we apply the following validation constraints prior to expert review.

\paragraph{Cross-national inconsistencies.}
Generated questions occasionally contained country-specific financial conventions (e.g., currencies, indices, or terminology). All such cases are standardized to U.S.-based financial settings.

\paragraph{Precision mismatch.}
In some cases, displayed values were rounded while computations used full precision. We align computational outputs with the displayed numerical precision.

\paragraph{Incomplete input specification.}
Some questions omitted variables required for computation. These cases are revised to include all necessary inputs.

\paragraph{Unit consistency.}
Currency symbols and units were inconsistently applied across questions and solutions. All templates are standardized to consistent units.

\paragraph{Non-informative steps.}
Certain generated solutions decomposed simple calculations into trivial substeps or omitted intermediate reasoning. These solutions are revised to reflect substantive reasoning steps.

\paragraph{Multiple targets.}
Some templates requested multiple output values, complicating evaluation. We constrain all templates to require a single target.

\subsection{Expert Review and Annotation Protocol}
\label{expertreview}

To further enhance data quality, we recruited ten financial experts to review all validated templates. The expert panel consists of seven graduate students in economics, finance, and related quantitative disciplines, and three industry professionals with experience in quantitative research, financial engineering, and risk management. Annotators were selected through an internal vetting process to ensure domain expertise and professional credibility. Demographic details are provided in \autoref{demography}.

\paragraph{Annotation Platform.}
We developed a Streamlit-based annotation platform to facilitate efficient expert review. Implementation details are provided in \autoref{platform}.

\paragraph{Pilot Study.}
Prior to full annotation, we conduct a pilot study in which 20 templates are reviewed by all annotators to calibrate evaluation standards. After calibration, all annotators agreed on the correctness of the pilot templates.

\paragraph{Main Annotation.}
The remaining 270 templates are randomly distributed among annotators, with each template reviewed by a single expert. Out of 290 total templates, 29 are identified as incorrect and subsequently revised by financial experts. Summary statistics of identified issues are reported in \autoref{templateissues}.

\section{Annotation and Quality Control}

\subsection{Financial Expert Demography}
\label{demography}

To ensure the reliability and domain robustness of our benchmark, all annotations were conducted by a diverse team of financial experts and advanced students with strong quantitative and economic backgrounds.
The annotators collectively represent three major categories: (1) industry professionals in quantitative research and financial engineering, (2) postgraduate students specializing in finance, economics, and auditing, and (3) experienced annotators trained in data labeling and financial analysis.

Several annotators have extensive industry experience across financial technology, quantitative research, and trading, with prior roles in investment banks, hedge funds, and fintech companies. 
Others are graduate students conducting research in finance, economics, and auditing, contributing academic rigor and theoretical grounding. 
Together, they bring complementary expertise that enhances both the practical and analytical aspects of our benchmark construction.

\paragraph{Summary.}
Our benchmark construction relies on a team of ten highly qualified annotators, including three industry professionals with prior experience in quantitative research or trading, and seven academic annotators who are graduate students in finance, economics, and auditing. This balanced composition, encompassing strong and diverse backgrounds in computer science, mathematics, statistics, and finance, ensures both professional authenticity and academic depth. Their combined expertise provides a robust foundation for high-quality, domain-consistent annotations, contributing to the overall reliability of \ourdataset. The following are the details for each of them.

\textit{Annotator A:} Currently pursuing a Ph.D. at a leading university in Asia, this annotator previously worked as a quantitative researcher at a fintech company, with experience across multiple financial markets including domestic equities, U.S. equities, Hong Kong equities, and cryptocurrencies. Their research focused on financial data generation, risk modeling, and trading strategies. They have also served as a research lead in risk management at a cryptocurrency investment fund. This blend of academic research and cross-market industry practice enhances the robustness and domain relevance of the benchmark annotations.

\textit{Annotator B:} A Master’s student at a leading university with a strong undergraduate background in finance. They previously interned in the equity financing division of a major securities firm, contributing practical insights into capital markets and investment banking.

\textit{Annotator C:} A Master’s student at a top institution, holding a bachelor’s degree in economics. Their training bridges theoretical economics and applied policy research, enriching the annotation process with domain-specific understanding.

\textit{Annotator D:} Holds a bachelor’s degree in economics and has received graduate admission offers from top international institutions. 

Their interdisciplinary background strengthens the dataset’s coverage of trade and international finance contexts.

\textit{Annotator E:} Holds a bachelor’s degree in economics, providing a solid foundation in macroeconomic theory and financial principles that supports reliable annotation and consistency across financial texts.

\textit{Annotator F:} A Master’s student at a well-known university specializing in auditing and intelligent systems, with a research focus on large language model evaluation and its applications in auditing. Their familiarity with both auditing and financial concepts supports the annotation of financial news and auditing benchmarks from a research-oriented perspective.

\textit{Annotator G:} A Master’s student at a university recognized for its auditing and financial programs, with strong grounding in auditing, financial analysis, and data quality control. Their prior participation in annotation projects ensures consistent standards for annotation accuracy.

\textit{Annotator H:} A quantitative analyst with an MSc-equivalent degree in financial technology from a top UK university. They have prior experience at major global financial institutions, focusing on stochastic modeling, risk management, and process automation. They also contribute to research on large language models in finance and are advancing toward professional certification in investment analysis.

\textit{Annotator I:} A quantitative researcher at a global investment firm with prior experience at quantitative research and technology companies. Their work spans cross-asset systematic strategies, portfolio optimization, and machine learning applications in trading. They also serve as a teaching assistant for a postgraduate course on systematic trading strategies.

\textit{Annotator J:} A quantitative trading analyst focused on equity derivatives, holding a postgraduate degree in financial engineering and risk management from a top European university and a bachelor’s degree from a globally recognized institution. Their professional experience includes roles at several financial institutions across asset management, banking, and fintech, covering alpha-signal development, portfolio optimization, and derivatives trading.

\subsection{Annotation Platform}
\label{platform}

We developed a custom annotation platform to evaluate the correctness of Python templates that generate financial questions and solutions. Each template corresponds to a financial scenario (e.g., investment analysis, compound interest, deposits, or ratio calculations). Annotators are instructed to review the code and determine whether both the financial framework and its implementation are correct, and whether the output representation (e.g., units, rounding) complies with the annotation policy.  

The annotation task requires a binary verdict: \textit{Correct} or \textit{Incorrect}. Templates labeled as \textit{Correct} need no modifications, though annotators may optionally provide comments. Templates labeled as \textit{Incorrect} must be associated with one or more issue tags, accompanied by a minimal code correction and a brief explanation.  
A template is considered \textit{Correct} when its financial framework, calculations, and representation fully conform to the policy. It is marked as \textit{Incorrect} if any substantive flaw is present in framework selection, mathematical logic, representation, robustness, or clarity. To facilitate consistent labeling, we introduced five issue tags:  
\begin{itemize}
    \item \textbf{Formula Choice Error}: An incorrect financial framework or formula is applied (e.g., simple vs.\ compound interest).  
    \item \textbf{Math/Logic Error}: Arithmetic or algorithmic errors within the chosen formula (e.g., $r \times n$ instead of $r / n$).  
    \item \textbf{Representation Error}: Inconsistent or incorrect handling of numbers, units, or rounding (e.g., annual vs.\ monthly mismatch).  
    \item \textbf{Robustness Error}: Failures on boundary or extreme inputs (e.g., division by zero, negative values).  
    \item \textbf{Clarity Issue}: Ambiguous variable names or comments that hinder auditability, even if the numerical results are correct.  
\end{itemize}  

\begin{figure*}[t]
    \centering
    \includegraphics[width=1.0\linewidth]{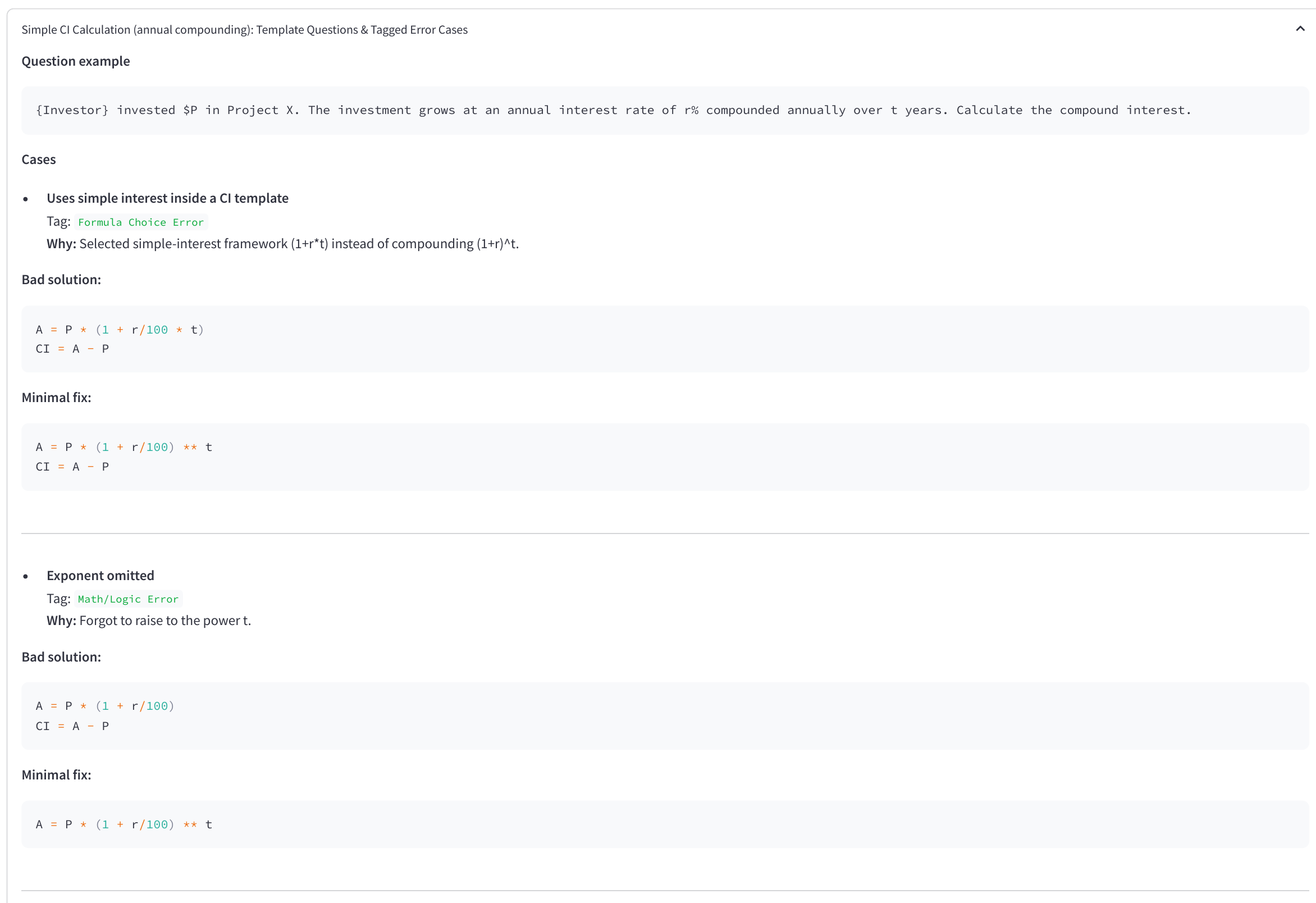}
    \caption{Reference examples for compound interest templates, illustrating typical annotation cases with error tags, flawed solutions, and minimal fixes.}
    \label{fig:compound-interest-examples}
\end{figure*}

To further support annotators, the platform provides curated reference cases across five templates within a single finance topic: compound interest. 
Each case includes (1) a question example, (2) a potential error type aligned with one of the defined issue tags, (3) a bad solution illustrating the error, and (4) a minimal code fix. 

\autoref{fig:compound-interest-examples} shows two representative cases: a \textit{Formula Choice Error}, where simple interest is incorrectly applied in a compound interest setting, and a \textit{Math/Logic Error}, where the exponent is omitted. Such examples provide concrete guidance for annotators, ensuring consistency and reliability.  

\begin{figure*}[t]
    \centering
    \includegraphics[width=1\linewidth]{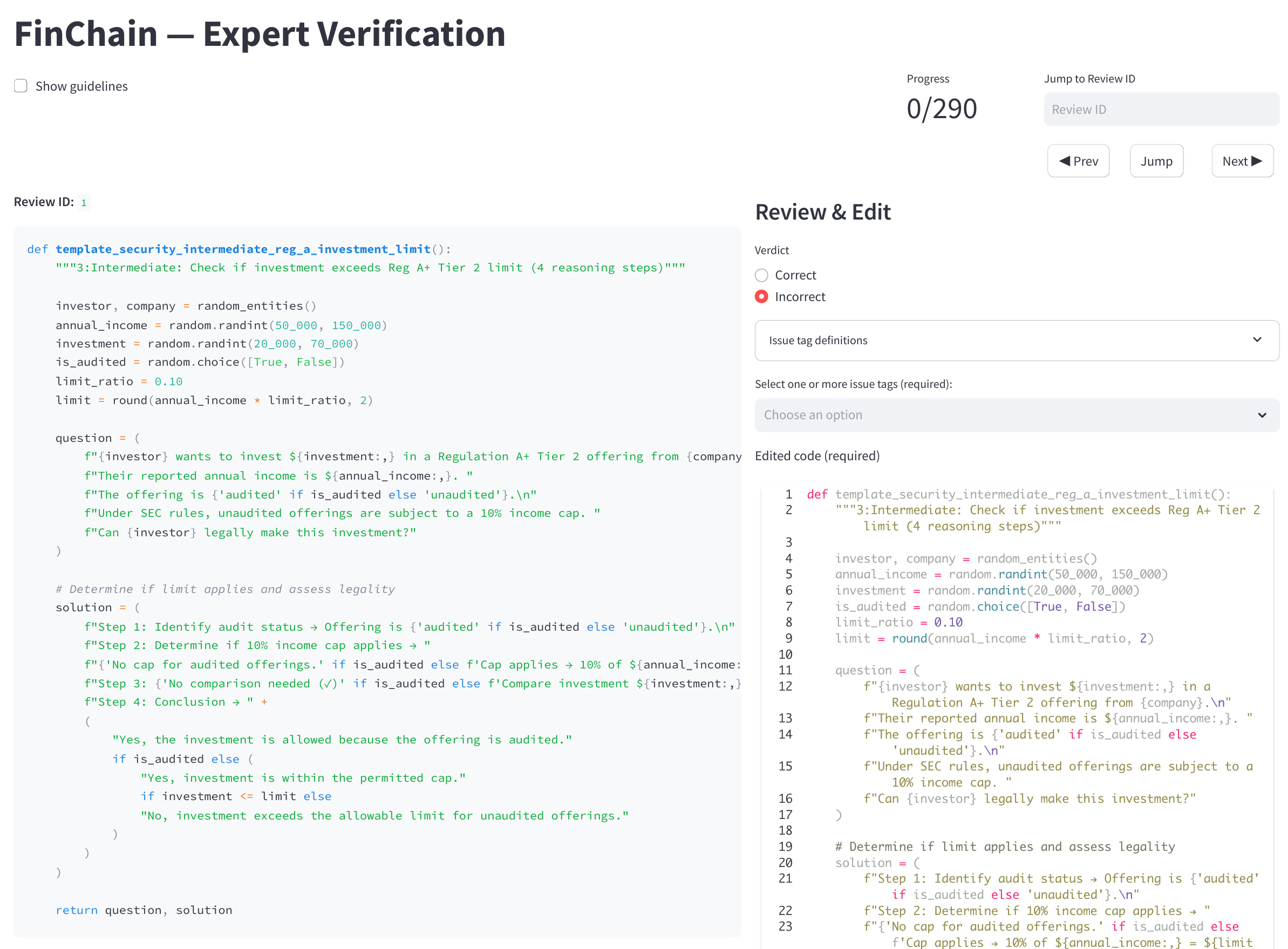}
    \caption{Expert annotation interface. Annotators review each template, assign a verdict, select issue tags, and provide minimal code corrections.}
    \label{fig:finchain-annotation}
\end{figure*}

After reviewing these reference cases, annotators proceed to the main annotation interface, where they evaluate unseen templates (\autoref{fig:finchain-annotation}). For each template, annotators must issue a binary verdict, select one or more issue tags if applicable, and provide a minimal code correction with a short justification. The interface presents the Python template and its generated question on the left, while the right panel allows annotators to record their verdict, choose tags, and edit the code directly. This design mirrors realistic auditing conditions and ensures that annotations capture both error identification and corrective reasoning.  

\subsection{Annotated Template Issue Statistics}
\label{templateissues}
Out of 290 templates, 29 (10\%) were tagged as containing errors during the annotation process. 
We summarize the distribution of issue types among annotated templates in~\autoref{tab:template-issues}. 

Most problems stem from representation and clarity errors, followed by formula selection, logical inconsistencies, and robustness issues.

\begin{table}[t]
\centering
\small
\begin{tabular}{lcc}
\toprule
\textbf{Issue Type} & \textbf{Count} & \textbf{Proportion (\%)} \\
\midrule
Representation Error & 12 & 41.4 \\
Clarity Issue         & 9  & 31.0 \\
Formula Choice Error  & 5  & 17.2 \\
Math/Logic Error      & 3  & 10.3 \\
Robustness Error      & 2  & 6.9 \\
\midrule
\textbf{Total Tagged Templates} & \textbf{29} & \textbf{100.0} \\
\bottomrule
\end{tabular}
\caption{Distribution of issue types among annotated templates.}
\label{tab:template-issues}
\end{table}

\subsection{Review Rubrics}
\label{rubrics}

To ensure fair and interpretable human evaluation, each model response is assessed along two complementary dimensions: \textit{Reasoning Process Quality} and \textit{Final Answer Accuracy}. For each question, reviewers are provided with the question itself, the standard reference answer, and the generated responses from different models. They independently assign scores on a 1–5 scale for each dimension, following the detailed rubrics below.

\subsection{Reasoning Process Quality}

This dimension evaluates how clearly, logically, and correctly the model articulates its reasoning steps leading to the final answer. High-quality reasoning should demonstrate coherent logical flow, factual correctness, and consistency with valid domain principles.

\begin{itemize}
    \item \textbf{1 (Unacceptable):} Illogical, incoherent, or irrelevant reasoning; missing steps or severe conceptual errors.
    \item \textbf{2 (Poor):} Some reasoning attempt but with major factual or procedural flaws; inconsistent or unclear Chain-of-Thought.
    \item \textbf{3 (Fair):} Partial understanding with mixed correct and incorrect reasoning; superficial or incomplete explanation.
    \item \textbf{4 (Good):} Mostly correct and coherent reasoning with minor inaccuracies or unclear phrasing; logical flow generally sound.
    \item \textbf{5 (Excellent):} Clear, well-structured, and logically consistent reasoning throughout; fully correct and well-justified steps.
\end{itemize}

\input{tables/model_notes}

\subsection{Final Answer Accuracy}

This dimension evaluates the correctness and completeness of the model’s final output relative to the reference solution. Reviewers compare each model’s final answer with the standard answer to determine whether the model’s conclusion is correct and sufficiently supported.

\begin{itemize}
    \item \textbf{1 (Unacceptable):} Completely incorrect or missing answer; no alignment with the reference solution.
    \item \textbf{2 (Poor):} Largely incorrect due to major conceptual or computational errors.
    \item \textbf{3 (Fair):} Partially correct; captures some relevant elements but omits or distorts key aspects of the correct solution.
    \item \textbf{4 (Good):} Largely correct and complete with only minor inaccuracies that do not affect the main result.
    \item \textbf{5 (Excellent):} Fully correct, precise, and complete; matches the reference solution exactly or with an equivalent formulation.
\end{itemize}

\section{Model Detail Information}
\label{modelcards}

\autoref{model-notes} provides details about the evaluated models.


\section{Metric Evaluation and Ablations}
\label{appx:metric_ablation}


In order to better understand the behavior of our proposed evaluation measure, we conducted a series of ablation experiments and comparative analysis.
All quantitative results reported in this section are benchmarked against expert human evaluations of reasoning quality (see \autoref{demography} for expert details), ensuring robustness and consistency across settings.

\subsection{Ablations of the DTW-Based Metric}
Our main evaluation metric, the \textbf{Normalized DTW Alignment Score (Gate Mode)}, measures both local semantic-numeric agreement and global sequence-level alignment between predicted and gold reasoning traces.  

To assess its robustness and the effect of its design choices, we considered several variants:

\begin{itemize}
    \item \textbf{DTW Gate Mode.}  
    This is the primary formulation used in the paper.
    Semantic similarity and numeric agreement are combined multiplicatively, i.e.,
    $\mathrm{Score}_{\mathrm{gate}}(i,j) = \mathrm{SS}(s^*_i, \hat{s}_j) \times \mathrm{AM}(s^*_i, \hat{s}_j)$.
    This ``gating'' ensures that steps contribute only when semantic meaning and intermediate results align consistently across diverse reasoning scenarios.    

    \item \textbf{DTW Soft Mode.}  
    A more permissive variant that blends semantic and numeric agreement through a weighted combination:
    $\mathrm{Score}_{\mathrm{soft}}(i,j) = 
    \alpha\, \mathrm{SS}(s^*_i, \hat{s}_j) + 
    \beta\, \mathrm{AM}(s^*_i, \hat{s}_j)$,  
    with $\alpha = 0.85$ and $\beta = 0.15$.  
    This ``soft'' formulation captures cases where partial numeric agreement still reflects correct reasoning, providing smoother sensitivity to small deviations. In other words, while the Gated version will assign 0 to a sequence of aligning reasoning steps, which resulted in a wrong answer (which can be a case if an LLM fails mathematics behind the solution), Soft version will still give a higher score.

    \item \textbf{DTW Precision, Recall, and F1.}  
    In addition to the normalized alignment score, we derive DTW-based precision, recall, and F1 measures that quantify step-level coverage under the DTW alignment path. These provide a finer breakdown of reasoning alignment.
\end{itemize}


To capture final correctness, we tried a weighted sum of DTW-based scores and final answer correctness. We varied $\alpha$ in the range $0.1$--$0.9$ to identify the best proportion. Based on these experiments, $\alpha = 0.1$ yielded the best Spearman $\rho$ and was used for comparison.

\subsection{Comparative Evaluation}
We also evaluated a range of traditional text-similarity and reasoning metrics, including ROUGE-2, ROUGE-L, step-level precision and recall (marked as `non-DTW' in the table), BERTScore, and our weighted sum of DTWNormGate and final answer's correctness (DTWNormGate+FAC in the table).  
Each metric was correlated with expert-assigned \emph{Reasoning Process Quality} score.
\autoref{tab:process_corr} summarizes the top Spearman correlations with expert process judgments.

\begin{table}[t]
\centering
\small
\begin{tabular}{lcc}
\toprule
\textbf{Metric} & \textbf{Spearman $\rho$} \\
\midrule
\textbf{DTWNormGate+FAC} & \textbf{0.655} \\
DTWNormGate & 0.640 \\
DTW Precision (Soft) & 0.625 \\
DTW Precision (Gate) & 0.622 \\
DTW F1 (Soft) & 0.619 \\
DTW F1 (Gate) & 0.618 \\
Step Precision (non-DTW) & 0.604 \\
DTWNormSoft  & 0.592 \\
DTW Recall (Gate) & 0.573 \\
Step Recall (non-DTW) & 0.570 \\
DTW Avg. Path Score (Gate) & 0.529 \\
DTW Avg. Path Score (Soft) & 0.526 \\
DTW Recall (Soft) & 0.512 \\
ROUGE-2 & 0.469 \\
ROUGE-L & 0.434 \\
BERTScore & 0.287 \\
\bottomrule
\end{tabular}
\caption{Spearman correlation with expert evaluation of the reasoning process.}
\label{tab:process_corr}
\end{table}


As additional validation, we also measure other correlation metrics: Kendall tau (which measures similarity of rankings) and Pearson correlation. As shown in \autoref{tab:process_corr_pearson} and \autoref{tab:process_corr_tau}, our suggested metric still holds high correlation position even with other correlation metrics across different evaluation settings and conditions.

\begin{table}[t]
\centering
\small
\begin{tabular}{lcc}
\toprule
\textbf{Metric} & \textbf{Pearson $\rho$} \\
\midrule
\textbf{DTWNormGate+FAC} & \textbf{0.584} \\
\textbf{DTW F1 (Soft)} & \textbf{0.584} \\
DTW Precision (Soft) & 0.578 \\
DTW F1 (Gate) & 0.568 \\
DTW Avg. Path Score (Gate) & 0.556 \\
DTW Recall (Gate) & 0.550 \\
Step Recall (non-DTW) & 0.549 \\
DTW Precision (Gate) & 0.548 \\
Step Precision (non-DTW) & 0.530 \\
DTWNormSoft  & 0.515 \\
DTW Avg. Path Score (Soft) & 0.515 \\
DTW Recall (Soft) & 0.511 \\
DTWNormGate & 0.492 \\

ROUGE-L & 0.421 \\
ROUGE-2 & 0.420 \\
BERTScore & 0.300 \\
\bottomrule
\end{tabular}
\caption{Pearson correlation with expert evaluation of the reasoning process.}
\label{tab:process_corr_pearson}
\end{table}

\begin{table}[t]
\centering
\small
\begin{tabular}{lcc}
\toprule
\textbf{Metric} & \textbf{Kendall $\tau$} \\
\midrule
\textbf{DTW Precision (Gate)} & \textbf{0.511} \\
\textbf{DTWNormGate+FAC} & 0.509 \\
Step Precision (non-DTW) & 0.505 \\
DTW F1 (Gate) & 0.503 \\
DTWNormGate & 0.503 \\
DTW Precision (Soft) & 0.494 \\
DTW F1 (Soft) & 0.487\\
Step Recall (non-DTW) & 0.477 \\
DTW Recall (Gate) & 0.463 \\
DTWNormSoft  & 0.460 \\
DTW Avg. Path Score (Gate) & 0.420 \\
DTW Avg. Path Score (Soft) & 0.401 \\
DTW Recall (Soft) & 0.389 \\
ROUGE-2 & 0.360 \\
ROUGE-L & 0.330 \\
BERTScore & 0.213 \\
\bottomrule
\end{tabular}
\caption{Kendall $\tau$ correlation with expert evaluation of the reasoning process.}
\label{tab:process_corr_tau}
\end{table}

\subsection{Sensitivity to the Numerical Tolerance $\epsilon$}
To assess the robustness of ChainEval to the numerical tolerance parameter $\epsilon$, we vary $\epsilon$ over a broad range and measure the Spearman correlation between the resulting metric scores and expert human judgments on reasoning quality. The results are shown in Table~\ref{tab:epsilon_sensitivity}.

\begin{table}[t]
\centering
\small
\begin{tabular}{lc}
\toprule
$\epsilon$ & Spearman $\rho$ \\
\midrule
0.00 & 0.54 \\
0.01 & 0.60 \\
0.03 & 0.60 \\
0.05 & 0.64 \\
0.15 & 0.61 \\
0.20 & 0.61 \\
0.50 & 0.60 \\
0.75 & 0.59 \\
\bottomrule
\end{tabular}
\caption{Sensitivity analysis of the numerical tolerance parameter $\epsilon$ in ChainEval. We report the Spearman correlation between metric scores and expert human judgments.}
\label{tab:epsilon_sensitivity}
\end{table}

The results show that the correlation remains relatively stable across a broad range of tolerance values. Performance peaks at $\epsilon = 0.05$, which is also consistent with commonly used materiality thresholds in financial auditing. Overall, the variation is modest, suggesting that the main conclusions are not sensitive to a specific threshold choice.

\subsection{Discussion}
The DTW-based variants consistently achieve the highest correlation with expert judgments, with the \textbf{Normalized DTW Alignment Score (Gate Mode) with FAC} emerging as the most reliable indicator of reasoning faithfulness.  
Non-FAC and the ``Soft'' variant yields slightly lower but still strong correlations, suggesting that the gating formulation better captures strict consistency, while the soft variant provides smoother sensitivity to near-correct reasoning.  
Compared to traditional metrics such as ROUGE or simple step-level precision and recall, DTW captures not only semantic similarity but also the structural coherence and numerical consistency of reasoning chains.  
These results highlight the usefulness of our proposed metric.

\subsection{Suggestions For Reproducibility}
Although our approach shows a strong correlation with human evaluations, it is not without limitations.
The way in which the reasoning steps are parsed from the model's output plays an important role in the quality estimation. In this work we combine both LLM-based parsing and parsing using regular expressions. We try to split the response on reasoning steps using regular expressions, which capture `Step X'-like patterns in the response, if such patterns are not found, we instruct LLM to split the answer on steps, copy-pasting the whole step string. We also use LLM to parse the final answer from the response. We acknowledge that such approaches depend on the prompting strategy, and there is a chance that other parsing and comparison methods (like the use of executable symbolic engine such as SymPy) can produce varying results.
For better reproducibility we share our parsing model details and prompts used to extract final answer and reasoning steps. 

We used the GPT-4.1 model. Results were parsed using OpenAI's output schema, which allowed us to avoid inconsistency in outputs. For the system prompt we used the following text:
\begin{quote}
\small
\begin{verbatim}
You are a strict financial reasoning parser.
Your task is to convert a noisy, long, 
conversational Chain-of-Thought solution 
into a clean and structured JSON object.
Follow these rules exactly:
1. No Calculations
Do NOT compute anything.
Do NOT recompute numbers from the text.
Do NOT round, simplify, 
or adjust any numeric value.
2. Number Extraction (CRITICAL)
You must copy numbers 
EXACTLY as they appear in the text.
Keep the sign intact 
(do NOT remove or change negative signs).
Keep decimal precision exactly as written.
Remove commas, dollar signs, 
and percent signs only as formatting, 
not as value changes.
3. Step Extraction
Identify only the actual reasoning steps.
Ignore filler text: summaries, 
meta-comments, confidence statements,
restatements, chatter.
Produce concise step descriptions 
capturing the logic of each step.
For each reasoning step, output an object with:
{
"index": <step_number>,
"text": "<short description of the step>",
"value": <numeric result of that step or null>
}
4. How to Determine value
If the step contains a 
computation result, copy the 
final numeric result in that step.
Usually this is the last 
number in the step (e.g., number after "=").
If the step has no numeric result, 
set "value": null.
5. Final Answer Extraction
"final_answer" must be copied exactly
as the last numeric value in the entire solution.
Apply the same numeric-copy rules as above.
6. JSON Output Format
Return ONLY a valid JSON object of the form:
{
"steps": [...],
"final_answer": <number>
}
7. Forbidden Behaviors
Do NOT change "-0.2636" into "2636.0".
Do NOT remove negative signs.
Do NOT round 1.514016 to 1.51.
Do NOT compute new numbers.
Do NOT invent numbers not present in the text.
Do NOT choose earlier numbers 
if a later number is clearly the result.
Accuracy is measured strictly.
Copy numbers exactly.
\end{verbatim}
\end{quote}
\newpage
For the user prompt we used:
\begin{quote}
\small
\begin{verbatim}
Here is the full solution text 
to parse into reasoning steps 
and a final answer:
f"{text}\n\n"
Remember: follow the instructions and 
return ONLY the JSON object.
\end{verbatim}
\end{quote}

In case the steps were parsed with regular expression correctly, the final answer was retrieved using the following prompt:

\begin{quote}
\small
\begin{verbatim}
You are given a full solution text.
Ignore step segmentation. 
Your ONLY job is to extract 
'final_answer' as the last numeric 
value in the solution,
following the same numeric-copy
rules from the system prompt.
Return ONLY a JSON object of the form:
"{ \"final_answer\": <number or null> }\"
Here is the solution text:
\end{verbatim}
\end{quote}

\section{Complementary Results}

\subsection{Results on a Sampled Subset}
\label{appendix_overall}

Due to the high inference cost of Grok~4~Heavy, we evaluate this model on a randomly sampled subset of 200 instances rather than the full benchmark.
The results in \autoref{tab:200sample} are reported to provide a coarse reference for this cost-limited setting and should not be directly compared with the full-benchmark results in \autoref{tab:overall_model_performance}.

\begin{table}[t]
\centering
\small
\begin{tabular}{l c c c}
\toprule
\textbf{Model} & \textbf{\ourmetric} & \textbf{FAC} & \textbf{ROUGE R$_2$} \\
\midrule
Grok 4 Heavy & 65.64 & \textbf{81.00} & 23.87 \\
GPT-5 & \textbf{68.42} & 69.50 & \textbf{28.37} \\
Gemini 2.5 Pro & 67.92 & 73.00 & 18.85 \\
Fin-R1 & 56.44 & 34.00 & 5.43 \\
Mathstral & 64.02 & 51.00 & 18.99 \\
\bottomrule
\end{tabular}
\caption{\textbf{Performance comparison on a randomly sampled 200-instance subset using \ourmetric{}, final answer correctness (FAC), and ROUGE R$_2$.} 
These results are reported for reference only and are not directly comparable to full-benchmark results.}
\label{tab:200sample}
\end{table}

\subsection{Complementary Domain-Level Results}
\label{appendix_domains}

\begin{figure}[t]
    \centering
    \includegraphics[width=1.0\columnwidth]{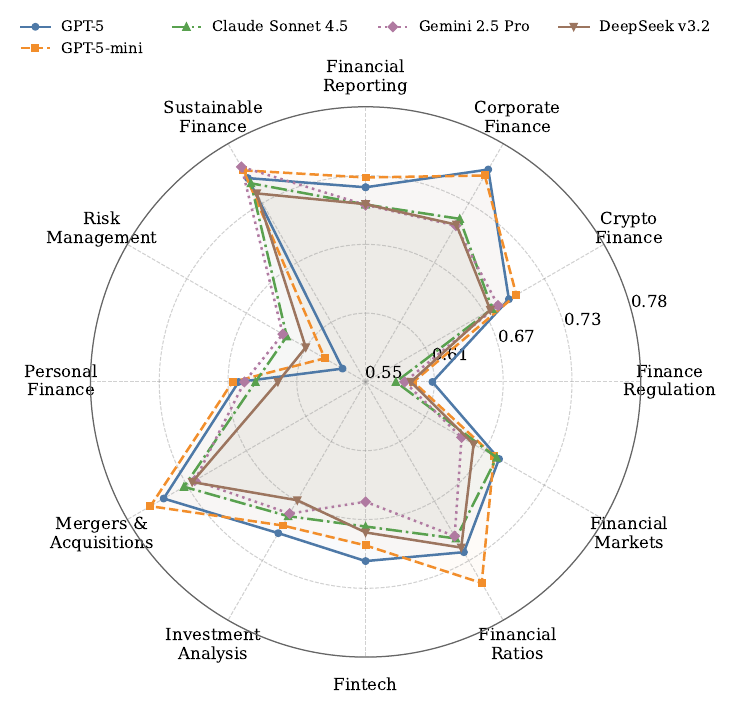}  
    \caption{
    \textbf{Domain-level performance of frontier proprietary models.}
Radar plot showing \ourmetric\ scores across twelve financial domains for GPT-5, GPT-5-mini, Claude Sonnet~4.5, Gemini~2.5~Pro, and DeepSeek~v3.2.
}
    \label{fig:topic_performance1}
\end{figure}

\begin{figure}[h!]
    \centering
    \includegraphics[width=1.0\columnwidth]{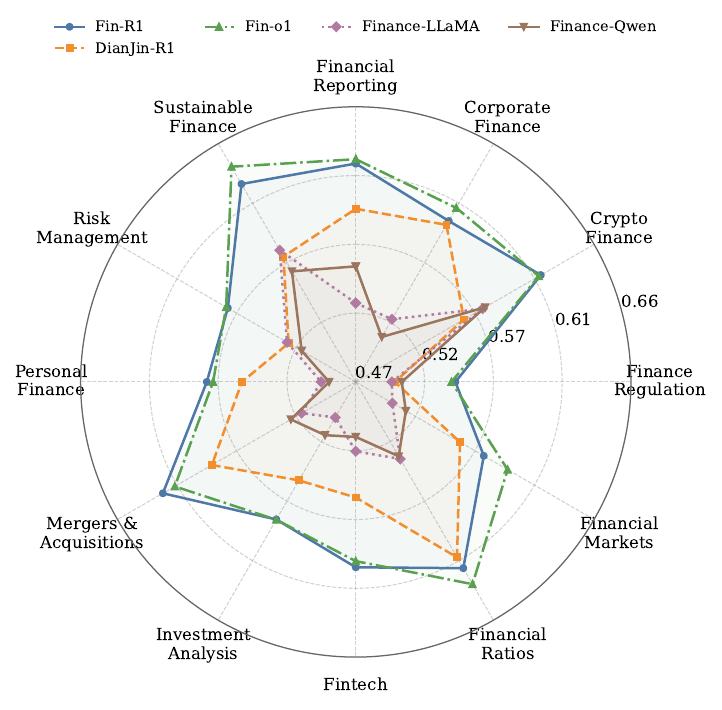}  
    \caption{
    \textbf{Domain-level performance of finance-tuned models.}
Radar plot showing \ourmetric\ scores across twelve financial domains for Fin-R1, DianJin-R1, Fin-o1, Finance-LLaMA, and Finance-Qwen.
}
    \label{fig:topic_performance2}
\end{figure}

We report additional domain-level results for the remaining models grouped by model category. These figures are provided for completeness and transparency, and no further analysis is conducted.
\autoref{fig:topic_performance1} reports domain-level \ourmetric\ scores for additional frontier proprietary models, showing broadly consistent performance patterns across financial domains with moderate variation.
\autoref{fig:topic_performance2} presents domain-level performance for finance-tuned models, illustrating heterogeneous patterns across domains.

\begin{figure}[t]
    \centering
    \includegraphics[width=1.0\columnwidth]{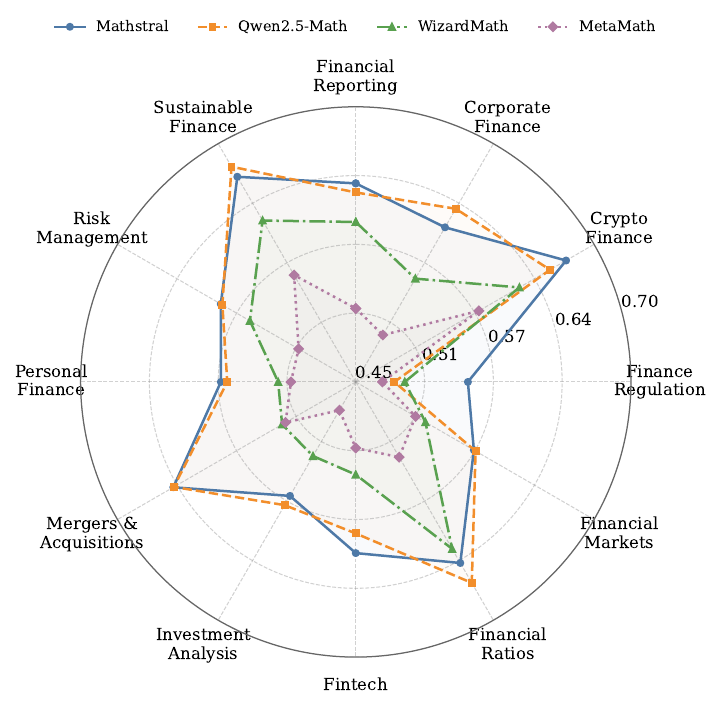}  
    \caption{
    \textbf{Domain-level performance of math-enhanced models.}
Radar plot showing \ourmetric\ scores across twelve financial domains for Mathstral, Qwen2.5-Math, WizardMath, and MetaMath.
}
    \label{fig:topic_performance3}
\end{figure}

\begin{figure}[t]
    \centering
    \includegraphics[width=1.0\columnwidth]{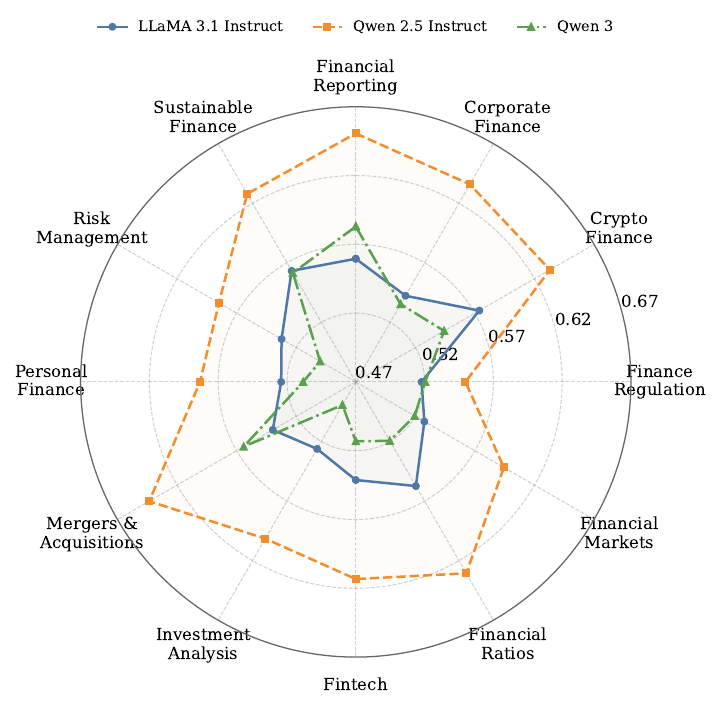}  
    \caption{
    \textbf{Domain-level performance of general-purpose open models.}
Radar plot showing \ourmetric\ scores across twelve financial domains for LLaMA~3.1~Instruct, Qwen~2.5~Instruct, and Qwen~3.
}
    \label{fig:topic_performance4}
\end{figure}

\autoref{fig:topic_performance3} shows domain-level results for math-enhanced models, with performance varying across financial domains.
\autoref{fig:topic_performance4} reports domain-level performance for general-purpose open models, serving as reference baselines across domains.

\begin{figure}[h!]
    \centering
    \includegraphics[width=1.0\columnwidth]{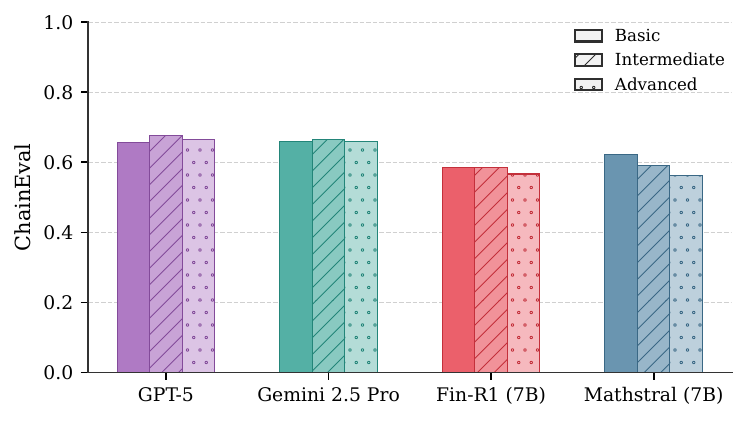}  
    \caption{\textbf{\ourmetric{} across difficulty levels.}
\ourmetric{} scores for representative models on \textit{Basic}, \textit{Intermediate}, and \textit{Advanced} \ourdataset instances. Compared to FAC in the main text, \ourmetric{} varies more gradually across difficulty, reflecting partial credit from step-level alignment.}
\label{fig:appx_chaineval_difficulty}
\end{figure}

\subsection{Difficulty Breakdown under \ourmetric{}}
\label{app_diff_chaineval}

As shown in \autoref{fig:appx_chaineval_difficulty}, \ourmetric{} exhibits a more gradual change across difficulty tiers than FAC. This indicates that models may maintain partial step-level alignment on harder instances even when end-task success decreases, consistent with \ourmetric{} assigning partial credit to intermediate reasoning.

\section{Error Analysis Supplementary}

\subsection{Error Taxonomy}
\label{appendix_error_taxonomy}

This appendix defines the error categories used in the diagnostic error analysis. Each model output is assigned to a single category based on the primary criterion violated in the predicted reasoning or final result.

\textbf{No Error.}
The reasoning process and final result match the gold computation within the predefined tolerance.

\textbf{Computational.}
The applied formula or method is appropriate, but one or more intermediate or final numerical computations are incorrect.

\textbf{Conceptual.}
The reasoning violates a problem constraint or applies a financial rule or formula inconsistently with the gold specification.

\textbf{Sign / Direction.}
The sign or directional definition of a quantity is inconsistent with the gold formulation, such as reversing subtraction order or treating a decrease as an increase.

\textbf{Unit / Precision.}
Units, scale, or numerical precision are handled inconsistently with the problem specification, for example percent versus decimal or dollars versus millions.

\textbf{Parsing / Format.}
The output structure is malformed or inconsistent, preventing reliable parsing or alignment of reasoning steps.

\textbf{Hallucination.}
One or more variables, assumptions, or quantities are introduced that are not supported by the original question or provided context.

\begin{table*}[t]
\centering
\small
\begin{tabular}{l l p{5.2cm} p{5.0cm}}
\toprule
\textbf{Error Type} & \textbf{Model} & \textbf{Model Output (snippet)} & \textbf{Error Description} \\
\midrule
Computational & GPT-5-mini &
``Total cash = 3,592,610.3 $\times$ 1.8632 = \$6,693,751.51.'' &
Incorrect multiplication; correct total is approximately \$6.68M. \\

Conceptual & GPT-5-mini &
``47408 + x = 0.8(60427 + x) $\Rightarrow$ x = 4668.'' &
Violates conservation of total portfolio value during rebalancing. \\

Sign / Direction & GPT-5-mini &
``Change = new - original $\approx$ -0.123.'' &
Uses an inconsistent sign convention; gold definition is old - new. \\

Unit / Precision & GPT-5-mini &
``175 (whole credits) ... total rebate = 175 $\times$ 2.03 = \$355.25.'' &
Applies unjustified rounding to a fractional quantity (175.5 credits). \\

Parsing / Format & Finance-Qwen &
``... on the investment at the end of 3 years?'' &
Output is malformed and cannot be reliably parsed into steps. \\

Hallucination & Finance-Qwen &
``Assume 500M shares ... Market cap = EPS $\times$ P/E = \$33.74B.'' &
Introduces an unsupported quantity and applies an inconsistent valuation formula. \\
\bottomrule
\end{tabular}
\caption{Representative error examples with model outputs (expert-audited sample).}
\label{tab:error_examples}
\end{table*}

\begin{table*}[t]
\centering
\small
\setlength{\tabcolsep}{4pt}
\begin{tabular}{lcccccc}
\toprule
\textbf{Error Type} & \textbf{GPT-5-mini} & \textbf{Finance-Qwen} & \textbf{DeepSeek v3.2} & \textbf{Fin-R1} & \textbf{Mathstral} & \textbf{Qwen 2.5 Instr.} \\
\midrule
No Error         & 80.5\% (161) & 10.0\% (20)  & 70.5\% (141) & 39.0\% (78)  & 23.5\% (47)  & 34.0\% (68)  \\
Computational    & 7.5\% (15)   & 24.5\% (49)  & 18.0\% (36)  & 35.5\% (71)  & 36.0\% (72)  & 44.0\% (88)  \\
Conceptual       & 9.5\% (19)   & 31.0\% (62)  & 8.5\% (17)   & 18.0\% (36)  & 30.5\% (61)  & 17.0\% (34)  \\
Sign/Direction   & 1.0\% (2)    & 0.0\% (0)    & 0.5\% (1)    & 0.5\% (1)    & 2.0\% (4)    & 0.5\% (1)    \\
Unit/Precision   & 1.5\% (3)    & 2.0\% (4)    & 1.5\% (3)    & 4.5\% (9)    & 6.0\% (12)   & 2.0\% (4)    \\
Parsing/Format   & 0.0\% (0)    & 15.5\% (31)  & 1.0\% (2)    & 1.0\% (2)    & 1.5\% (3)    & 1.0\% (2)    \\
Hallucination    & 0.0\% (0)    & 17.0\% (34)  & 0.0\% (0)    & 1.5\% (3)    & 0.5\% (1)    & 1.5\% (3)    \\
Error Rate       & 19.5\% (39)  & 90.0\% (180) & 29.5\% (59)  & 61.0\% (122) & 76.5\% (153) & 66.0\% (132) \\
\bottomrule
\end{tabular}
\caption{Expanded error analysis across six models on the same sampled set of 200 questions.}
\label{tab:error_analysis_6models}
\end{table*}

\newpage
\subsection{Error Examples}
\label{appendix_error_examples}

\autoref{tab:error_examples} presents one representative example for each error category used in the diagnostic analysis.
Each example consists of a short excerpt from the model output and a concise description of the specific criterion under which the output is labeled as erroneous. The descriptions focus on observable mismatches between the model output and the corresponding gold computation or definition, such as numerical inconsistency, violation of problem constraints, or unsupported quantities.
The examples are intended solely to illustrate how the error taxonomy is applied in practice. 

They do not aim to explain the underlying causes of model behavior, assess error frequency, or attribute failures to model training, architecture, or reasoning capacity.

\subsection{Expanded Error Analysis}
\label{appx:error_analysis_extended}


To test whether the qualitative findings from the main-text error analysis generalize beyond two representative systems, we extend the same annotation protocol to six models spanning different performance levels: \texttt{GPT-5-mini}, \texttt{Finance-Qwen}, \texttt{DeepSeek v3.2}, \texttt{Fin-R1}, \texttt{Mathstral}, and \texttt{Qwen 2.5 Instruct}. All models were evaluated on the same randomly sampled subset of 200 questions, ensuring a controlled and directly comparable evaluation setting across systems.

The full results are presented in \autoref{tab:error_analysis_6models}, which provides a detailed breakdown of error categories for each model.

The expanded comparison yields the same qualitative pattern as in the main text. Stronger models tend to fail primarily on computational and conceptual mistakes, typically arising from arithmetic inaccuracies or incorrect application of financial formulas, whereas lower-performing models exhibit a broader and more heterogeneous error profile, including substantially more parsing/format issues and hallucinated reasoning steps that are unsupported by the input.

Overall, the expanded analysis strengthens the robustness of our qualitative findings and shows that the observed error patterns are not limited to two representative models, but generalize consistently across models with varying performance levels and training paradigms.

%% file: tables/example1.tex
\begin{lstlisting}[style=templcode]
def template_ci_quarterly_compounding():
    """Basic: Compound Interest with Quarterly Compounding"""
    investor_name = random.choice(investor_names)
    project_name  = random.choice(project_names)

    # Parameters
    principal = random.randint(1_000, 7_000)       # $
    rate      = round(random.uniform(2, 8), 2)     # annual %, two decimals
    time      = random.randint(1, 3)               # years
    n         = 4                                   # quarterly

    # ---------- Question ----------
    question = (
        f"{investor_name} invests ${principal} in {project_name}. "
        f"The account earns {rate:.2f}% interest per year, compounded quarterly, "
        f"for {time} years. What is the total compound interest earned "
    )

    # ---------- Reasoning ----------
    # Step 1: future (compound) amount
    future_value = principal * (1 + rate / (100 * n)) ** (n * time)
    # Step 2: compound interest
    ci = future_value - principal

    # Round only for display
    fv_display = f"${future_value:.2f}"
    ci_display = f"${ci:.2f}"

    # ---------- Solution ----------
    solution = (
        "Step 1. Compute the future value with quarterly compounding:\n"
        "  n = 4 periods per year.\n"
        "  Future Value = P × (1 + r / (100 × n))^(n × t)\n"
        f"              = ${principal} × (1 + {rate:.2f}% / (100 × 4))^(4 × {time})\n"
        f"              = ${principal} × (1 + {rate / (100 * n):.4f})^{4*time}\n"
        f"              = {fv_display}\n\n"
        "Step 2. Find the compound interest earned:\n"
        "  Compound Interest = Future Value − Principal\n"
        f"                   = {fv_display} − ${principal}\n"
        f"                   = {ci_display}"
    )

    return question, solution
\end{lstlisting}

%% file: tables/example2.tex
\begin{lstlisting}[style=templcode]
def template_ci_rate_and_total_known():
    """Intermediate: Compound Interest with nominal rate, time, and frequency known"""

    investor_name = random.choice(investor_names)
    project_name  = random.choice(project_names)

    # ---------- Parameters ----------
    total_amount = random.randint(5_000, 15_000)           # Final amount A ($)
    rate         = round(random.uniform(2, 10), 2)         # Nominal annual rate %
    time         = random.randint(1, 5)                    # Years
    freq_name, n = random.choice(
        [("semi-annually", 2), ("quarterly", 4), ("monthly", 12)]
    )

    # ---------- Question ----------
    question = (
        f"{investor_name} received a total amount of ${total_amount:,.2f} "
        f"from their investment in {project_name}. "
        f"The investment grew at a nominal annual interest rate of {rate:.2f}% "
        f"compounded {freq_name} for {time} years. "
        f"Calculate the compound interest earned (in dollars)."
    )
    # ---------- Reasoning ----------
    # Step 1: periodic rate and growth factor
    periodic_rate = round(rate / 100 / n, 6)               # r_p
    growth_factor = round((1 + periodic_rate) ** (n * time), 6)
    # Step 2: principal P
    principal = round(total_amount / growth_factor, 2)     # 2 dp dollars
    # Step 3: compound interest CI
    ci = round(total_amount - principal, 2)
    # ---------- Solution ----------
    solution = (
        "Step 1 – Find the periodic rate and growth factor\n"
        f"  Periodic rate  = {rate:.2f}% ÷ {n} = {periodic_rate*100:.4f}%\n"
        f"  Growth factor  = (1 + {periodic_rate:.6f})^{n*time} = {growth_factor:.6f}\n\n"
        "Step 2 – Compute the initial principal\n"
        f"  P = A ÷ growth factor = "
        f"${total_amount:,.2f} ÷ {growth_factor:.6f} = ${principal:,.2f}\n\n"
        "Step 3 – Calculate the compound interest\n"
        f"  CI = A − P = ${total_amount:,.2f} − ${principal:,.2f} = ${ci:,.2f}"
    )

    return question, solution
\end{lstlisting}

%% file: tables/example3.tex
\begin{lstlisting}[style=templcode]
def template_ci_with_additional_deposit():
    """Advanced: Compound Interest with a Mid Term Additional Deposit (needs 4 steps)"""
    investor_name = random.choice(investor_names)
    project_name  = random.choice(project_names)

    # --- parameters ---
    principal = random.randint(2000, 8000)          # initial $
    rate      = round(random.uniform(3, 10), 2)     # % p.a.
    time      = random.randint(3, 7)                # total years (>=3 so a mid deposit makes sense)
    n         = random.choice([1, 2, 4, 12])        # compounds per year

    deposit       = random.randint(500, 4000)       # extra $
    deposit_time  = random.randint(1, time - 1)     # year when deposit is made

    # ---------- Question ----------
    question = (
        f"{investor_name} initially invested ${principal} in {project_name} at an annual "
        f"rate of {rate:.2f}%, compounded {n} times a year, for a total of {time} years. "
        f"Exactly {deposit_time} years after the start, they added an extra ${deposit} "
        f"to the same account under the same rate and compounding frequency. "
        f"Calculate the total compound interest earned by the end of the {time} years."
    )

    # ---------- Reasoning ----------
    # Step 1 – periodic rate
    periodic_rate = round(rate / (100 * n), 4)

    # Step 2 – grow the original principal for the full period
    periods_principal = n * time
    fv_principal = round(principal * (1 + periodic_rate) ** periods_principal, 2)

    # Step 3 – grow the later deposit for the remaining (time - deposit_time) years
    remaining_years = time - deposit_time
    periods_deposit = n * remaining_years
    fv_deposit = round(deposit * (1 + periodic_rate) ** periods_deposit, 2)

    # Step 4 – combine amounts and find compound interest
    total_future_value   = round(fv_principal + fv_deposit, 2)
    total_contributions  = principal + deposit
    compound_interest    = round(total_future_value - total_contributions, 2)

    # ---------- Solution ----------
    solution = ("Step 1 – Periodic rate:\n"
        f"  r = {rate:.2f}% / (100 x {n}) = {periodic_rate:.4f}\n\n"
        "Step 2 – Future value of the original principal:\n"
        f"  Periods = {n} x {time} = {periods_principal}\n"
        f"  FV1 = ${principal} x (1 + {periodic_rate:.4f})^{periods_principal} = "
        f"${fv_principal:.2f}\n\n"
        "Step 3 – Future value of the additional deposit:\n"
        f"  Remaining years = {time} - {deposit_time} = {remaining_years}\n"
        f"  Periods = {n} x {remaining_years} = {periods_deposit}\n"
        f"  FV2 = ${deposit} x (1 + {periodic_rate:.4f})^{periods_deposit} = "
        f"${fv_deposit:.2f}\n\n"
        "Step 4 – Total compound interest:\n"
        f"  Total FV = FV1 + FV2 = ${fv_principal:.2f} + ${fv_deposit:.2f} = "
        f"${total_future_value:.2f}\n"
        f"  Contributions = ${principal} + ${deposit} = ${total_contributions}\n"
        f"  Compound Interest = Total FV - Contributions = "
        f"${total_future_value:.2f} - ${total_contributions} = ${compound_interest:.2f}")

    return question, solution
\end{lstlisting}

%% file: tables/model_notes.tex
\begin{table*}[t]
\centering
\resizebox{0.99\textwidth}{!}{%
\renewcommand{\arraystretch}{1.1}
\begin{tabular}{llllp{9cm}}
\toprule
Model & Organization & Size & Backbone & Source \\
\noalign{\vskip 0.5ex}\hdashline\noalign{\vskip 0.5ex}
\multicolumn{5}{l}{\textbf{Frontier Proprietary LLMs}} \\
\noalign{\vskip 0.5ex}\hdashline\noalign{\vskip 0.5ex}
GPT-5 & OpenAI & N/A & -- & \texttt{gpt-5-2025-08-07} \\
GPT-4.1 & OpenAI & N/A & -- & \texttt{gpt-4.1-2025-04-14} \\
GPT-5-mini & OpenAI & N/A & -- & \texttt{gpt-5-mini-2025-08-07} \\
GPT-4.1-mini & OpenAI & N/A & -- & \texttt{gpt-4.1-mini-2025-04-14} \\
Claude Sonnet 4.5 & Anthropic & N/A & -- & \texttt{claude-sonnet-4-5-20250929} \\
Claude Sonnet 4 & Anthropic & N/A & -- & \texttt{claude-sonnet-4-20250514} \\
Claude Sonnet 3.7 & Anthropic & N/A & -- & \texttt{claude-3-7-sonnet-20250219} \\
Gemini-2.5 Pro & Google & N/A & -- & \texttt{Last Update: June 2025} \\
Gemini-2.5 Flash & Google & N/A & -- & \texttt{Last Update: June 2025} \\
DeepSeek-V3.2 & DeepSeek & N/A & -- & \texttt{Last Update: Sep 29 2025} \\
DeepSeek-V3.1 & DeepSeek & N/A & -- & \texttt{Last Update: Sep 22 2025} \\
DeepSeek-R1 & DeepSeek & N/A & -- & \texttt{Last Update: Jan 20 2025} \\
Grok-4 Heavy & xAI & N/A & -- & \texttt{grok-4-0709} \\
Grok-4 Fast & xAI & N/A & -- & \texttt{grok-4-fast-reasoning} \\
\noalign{\vskip 0.5ex}\hdashline\noalign{\vskip 0.5ex}
\multicolumn{5}{l}{\textbf{Finance Specific LLMs}} \\
\noalign{\vskip 0.5ex}\hdashline\noalign{\vskip 0.5ex}
Fin-o1 & TheFinAI & 8B & \texttt{meta-llama/Llama-3.1-8B} & \texttt{TheFinAI/Fin-o1-8B} \\
Fin-R1 & SUFE-AIFLM-Lab & 7B & \texttt{Qwen/Qwen2.5-7B-Instruct} & \texttt{SUFE-AIFLM-Lab/Fin-R1} \\
DianJin-R1 & Qwen DianJin Team & 7B & \texttt{Qwen/Qwen2.5-7B-Instruct} & \texttt{DianJin/DianJin-R1-7B} \\
Finance-LLaMA & Wiro AI & 8B & \texttt{deepseek-ai/DeepSeek-R1-Distill-Llama-8B} & \texttt{WiroAI/WiroAI-Finance-Llama-8B} \\
Finance-Qwen & Wiro AI & 7B & \texttt{Qwen/Qwen2.5-7B} & \texttt{WiroAI/WiroAI-Finance-Qwen-7B} \\
\noalign{\vskip 0.5ex}\hdashline\noalign{\vskip 0.5ex}
\multicolumn{5}{l}{\textbf{Math Enhanced LLMs}} \\
\noalign{\vskip 0.5ex}\hdashline\noalign{\vskip 0.5ex}
WizardMath & WizardLM Team & 7B & \texttt{mistralai/Mistral-7B-v0.1} & \texttt{WizardLMTeam/WizardMath-7B-V1.1} \\
MetaMath & MetaMath Project & 7B & \texttt{EleutherAI/llemma 7b} & \texttt{meta-math/MetaMath-7B-V1.0} \\
Mathstral & Mistral AI & 7B & \texttt{mistralai/Mistral-7B-v0.1} & \texttt{mistralai/Mathstral-7B-v0.1} \\
Qwen-2.5-Math & Qwen Team & 7B & \texttt{Qwen/Qwen2.5-7B} & \texttt{Qwen/Qwen2.5-Math-7B-Instruct} \\
\noalign{\vskip 0.5ex}\hdashline\noalign{\vskip 0.5ex}
\multicolumn{5}{l}{\textbf{General Purpose Open LLMs}} \\
\noalign{\vskip 0.5ex}\hdashline\noalign{\vskip 0.5ex}
LLaMA-3.1 & Meta & 8B & - & \texttt{meta-llama/Llama-3.1-8B} \\
Qwen-2.5 & Qwen Team & 7B & - & \texttt{Qwen/Qwen2.5-7B-Instruct} \\
Qwen-3 & Qwen Team & 8B & - & \texttt{Qwen/Qwen3-8B} \\
\bottomrule
\end{tabular}}
\caption{Details of the organization and model source (\ie\ model version for proprietary models, and HuggingFace model name for open-source models) for the LLMs evaluated in \ourdataset.}
\label{model-notes}
\end{table*}